\documentclass[runningheads]{llncs}

\usepackage{eccv}

\usepackage{eccvabbrv}

\usepackage{graphicx}
\usepackage{booktabs}

\usepackage[accsupp]{axessibility}

\usepackage{hyperref}

\usepackage{orcidlink}

\usepackage{physics}
\usepackage{amsmath}
\usepackage{bm}
\usepackage[dvipsnames]{xcolor}
\usepackage[linesnumbered,ruled]{algorithm2e}

\SetCommentSty{mycommfont}
\SetKwInput{KwInput}{Input}                
\SetKwInput{KwOutput}{Output}

\def \bcB{{\beta^B}\xspace}
\def \bcD{{\beta^D}\xspace}

\def \bcB{{\phi_b}\xspace}
\def \bcD{{\phi_a}\xspace}
\def \Binf{{\phi}^\infty\xspace}

\newif\ifarxiv \newcommand{\arxiv}{\arxivtrue}
\newif\ifcamera 
\arxiv 

\begin{document}


\title{Osmosis: RGBD Diffusion Prior for Underwater Image Restoration} 

\titlerunning{Osmosis: RGBD Diffusion Prior for Underwater Image Restoration}

\author{Opher Bar Nathan\inst{1}\orcidlink{0009-0006-6013-2306} \and Deborah Levy\inst{1}\orcidlink{0000-0003-0080-1054} \and Tali Treibitz\inst{1}\orcidlink{0000-0002-3078-282X} \and Dan~Rosenbaum\inst{2}\orcidlink{0009-0008-9558-3195}}

\authorrunning{O.~Bar Nathan et al.}

\institute{Hatter Department of Marine Technologies, Charney School of Marine Sciences \and Department of Computer Science\\ University of Haifa, Haifa, Israel\\ 
\texttt{\href{https://osmosis-diffusion.github.io/}{\textcolor{magenta}{osmosis-diffusion.github.io}}}}

\maketitle

 \begin{abstract}
Underwater image restoration is a challenging task because of water effects that increase dramatically with distance. This is worsened by lack of ground truth data of clean scenes without water. Diffusion priors have emerged as strong image restoration priors. However, they are often trained with a dataset of the desired restored output, which is not available in our case.
We also observe that using only color data is insufficient, and therefore augment the prior with a depth channel. We train an unconditional diffusion model prior on the joint space of color and depth, using standard RGBD datasets of natural outdoor scenes in air. Using this prior together with a novel guidance method based on the underwater image formation model, we generate posterior samples of clean images, removing the water effects. 
Even though our prior did not see any underwater images during training, our method outperforms state-of-the-art baselines for image restoration on very challenging scenes.
Our code, models and data are available on the project’s website.


\end{abstract}    
\section{Introduction}
\label{sections:Introduction}

\begin{figure}[t]
    \centering
    \includegraphics[width=0.99\linewidth]{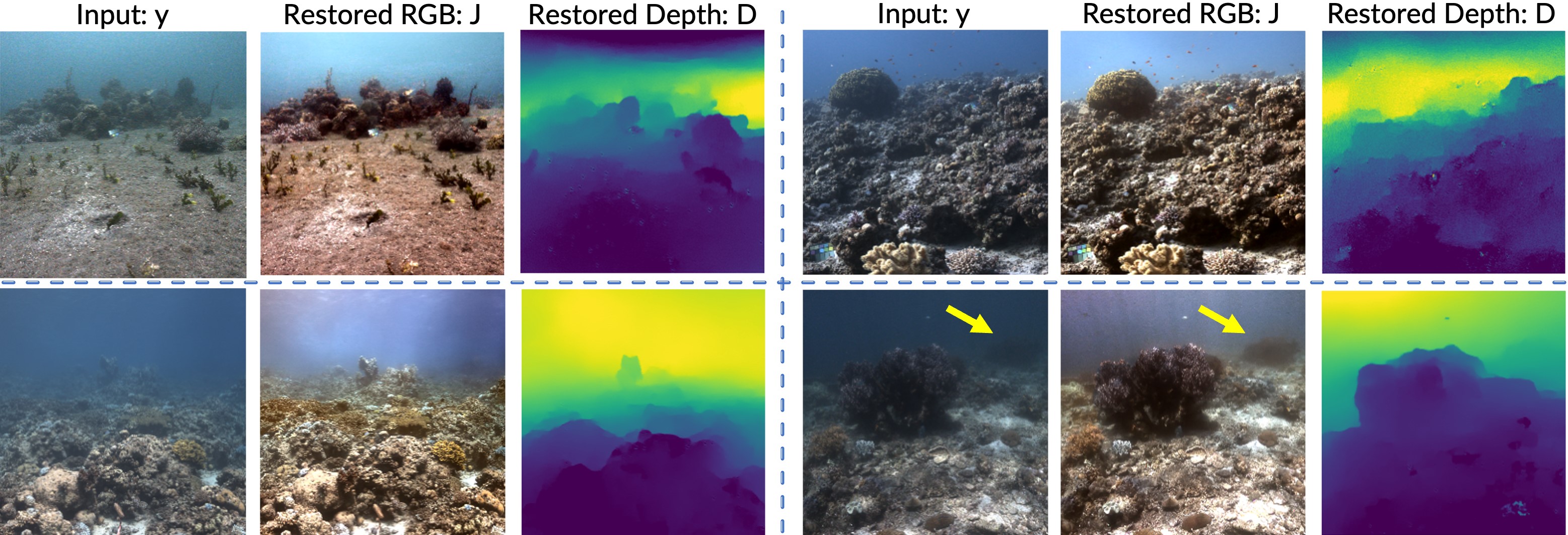}
    \caption{\textbf{Our method} receives as an input a single underwater image and outputs the restored clean image and an estimated depth map. The output is estimated using a diffusion prior trained on RGBD images and the physical image formation model.}
    \label{fig:teaser}
\end{figure}

\begin{figure}[t]
    \centering
    \includegraphics[width=0.96\linewidth]{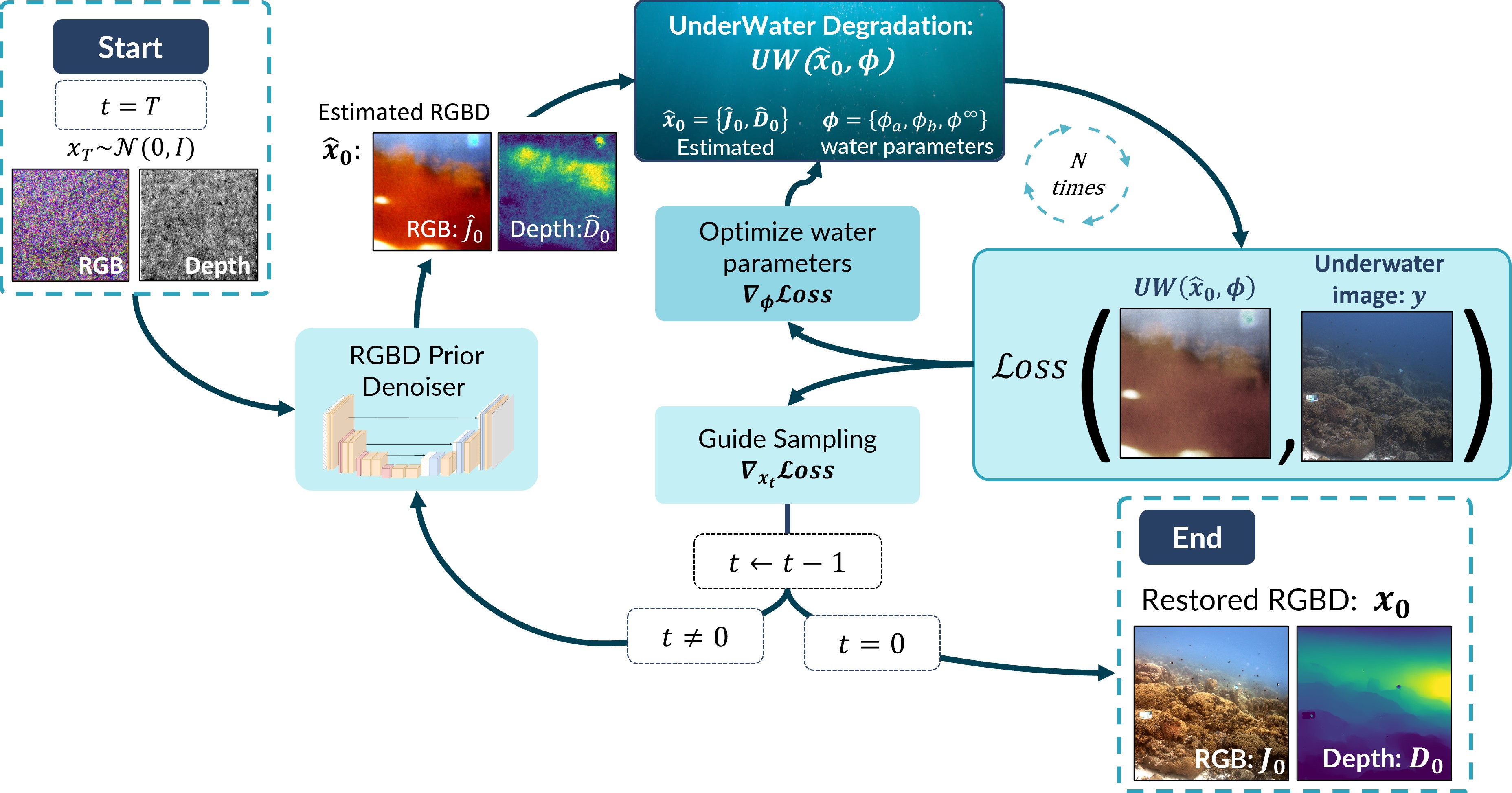}
    \caption{\textbf{The iterative sampling process} starts in $t=T$ with random noise in $4$ channels. The denoising step outputs
    denoised samples $\hat{x}_0= (\hat{J}_0, \hat{D}_0)$. We use the underwater physical image formation model together with $\hat{x}_0$ to optimize the water parameters $\hat{\phi}$, and to guide the sampling towards the observed image. This process repeats itself, gradually updating both the estimated image and depth, until $t=0$, in which $x_0$ holds the method's estimate for both the reconstructed scene $J_0$ and its depth $D_0$. }
    \label{fig:algoprocess}
\end{figure}

Underwater images are used in many applications, such as, underwater construction and maintenance, marine sciences, and fisheries. However, their automatic analysis is hindered because of the optical effects of the water that strongly attenuates and scatters light in a wavelength dependent manner. This causes color distortion and loss of contrast that exponentially increase with  depth\footnote{For consistency with computer vision literature the term \emph{depth} is used throughout to refer to the distance from the camera rather than   water depth.}.  With  growing human activity in the oceans, clear underwater vision becomes increasingly important.

Restoring underwater scenes is still a very challenging ill-posed problem. Classic approaches are based on designing priors for clean images or inverting the water formation model, and are limited by the ability to form strong priors.  On the other hand, learning based approaches are limited by the lack of supervised training data of clean underwater images. This is a critical issue, as the ocean cannot be emptied for the sake of data collection. 

We suggest an \emph{unsupervised} restoration method (Figs.~\ref{fig:teaser},\ref{fig:algoprocess}) based on an \emph{inverse problem} approach using a diffusion prior for both color and depth, coupled with the underwater image formation model. Restoring an image is formulated as posterior inference, computed using a  natural image prior, and a likelihood term that is based on the underwater image formation model. 
The challenge with applying this approach for underwater image restoration directly is that (i) the degradation for each pixel depends on its depth, and other unknown parameters; and (ii) there is no ground truth clean underwater data to train the prior.

To solve this, we replace the image prior with a prior on the joint space of color and depth of natural images. Adding depth to the prior allows us to formulate the forward model that forms the likelihood term over the observed corrupted underwater image, and apply posterior sampling. Moreover, this leverages the high capacity of diffusion models, in capturing the strong correlations between color and depth in natural scenes. 

We propose to train a prior model of RGBD images using available datasets of natural outdoor scenes that were collected in air. Using in-air scenes for underwater images may be counter-intuitive, but actually has strong benefits: 1)~it overcomes the lack of clean underwater image data; 2)~it leads to a strong prior that captures the joint statistics governing color and depth in natural scenes, where, as opposed to underwater images, color does not fade with distance; 3)~it prevents overfitting to specific types of underwater images.

We then use this prior together with the underwater physical image formation model to simultaneously estimate the clean image, its depth, and the model parameters, all from a single underwater image (summarized in Fig.~\ref{fig:algoprocess}). We show that our method outperforms models that were trained on underwater data. 

\newpage
\noindent\textbf{The main contributions of this paper are:} 

\textbf{1.} We train an RGBD prior, and demonstrate that modeling color and depth and jointly sampling them, provides a stronger diffusion prior for underwater image restoration.

\textbf{2.} We propose a new method that combines the RGBD prior of in-air data with the underwater image formation model, leading to a diffusion guidance method that generates the restored underwater images as posterior samples. 

\textbf{3.} We demonstrate that our method outperforms state-of-the-art underwater image restoration methods both qualitatively and quantitatively on real and simulated data. 

\noindent We publish all our code, the new trained RGBD prior, images and results. 

\section{Related Works}
\label{sections:related_work}

\subsection{Underwater Image Restoration and Depth Estimation}

Here we review recent works that are most relevant to our method. See~\cite{zhou2023underwater} for a recent more comprehensive review.

\noindent\textbf{Classic underwater image restoration.}
To cope with the ill-posed nature of the problem, earlier works  introduced tailored image priors. Some priors aim to estimate a depth or transmission map of the scene to reduce the number of the unknowns and then use the estimated depth to restore the scenes~\cite{peng2017underwater,peng2018generalization,berman2020underwater}. 

\noindent\textbf{Supervised learning.} Clean underwater image data is very scarce. 
Several datasets aim to mitigate this issue. 
In UIEB~\cite{li2019underwater}, the images denoted as ground-truth are generated from enhancement results of 9 different baseline methods, and having human observers vote for the best one. The LSUI dataset~\cite{peng2023u} is larger, produced using the same methodology, based on choosing results from 18 different methods. These datasets enable supervised learning methods, but are still limited by their scale, the quality of the baseline methods, and the human bias to choose visually pleasing rather than physically consistent images.

CWR~\cite{han2022underwater} introduced the HICRD dataset, where the images are restored using optical parameters impressively measured using ocean optics instruments.  Unfortunately, the images are acquired in a downward-looking position, and thus their depth range is very limited. In FUnIE-GAN~\cite{islam2020fast} a training dataset is generated by having humans select  \emph{good}  images from a large set of unprocessed underwater images. The humans are instructed to choose images where the \emph{foreground} objects are identifiable. These are then distorted by a GAN to produce the paired \emph{poor} images. 
A synthetic dataset was generated in~\cite{li2020underwater}. The dataset is synthesized from the NYU-v2 RGBD dataset~\cite{silberman2012indoor}, using the image formation model equation and several sets of values for the water parameters. 

\noindent\textbf{Unsupervised learning.} USUIR~\cite{fu2022unsupervised} aims to restore images without supervision, by separately estimating the image components (clean image, transmission, backscatter) and using them to construct an underwater image that is used for supervision against the original one. Subsequent frames were used in~\cite{amitai2023self} for self-supervising monocular depth estimation, and in~\cite{varghese2023self} for self-supervising both depth and restoration. In UW-NET~\cite{gupta2019unsupervised} a cycle-GAN is used for learning mapping from RGBD in-air datasets to underwater images. \textbf{As opposed to all these methods, we present the first prior based on diffusion models that does not rely on ground truth underwater supervision, and uses the physical model for inference.}

\subsection{Diffusion Model Prior}

\noindent\textbf{Diffusion models.}
Diffusion models have emerged as a powerful type of generative models. In the last few years several formulations and variations have been developed~\cite{sohl2015deep, song2019generative, ho2020denoising, song2020score}, most of which use a U-Net architecture~\cite{ronneberger2015u} as a noise predictor.  Because the training relies on very large datasets and is extremely time-consuming, significant work has been devoted to the setup where models are first pre-trained on large datasets and only later fine-tuned to more specific data, closer to the tasks at hand~\cite{ruiz2023dreambooth, sohn2023styledrop}. In DepthGen~\cite{saxena2023monocular} this approach is taken forward, by using a pre-trained model to kickstart a new model trained on different modalities using a different architecture. This is done by replacing the input and output layers of the pre-trained U-Net. We take a similar approach for training our RGBD prior, by using a pretrained diffusion model that was trained on RGB only.

Conditional diffusion models have been used for various image restoration tasks, by training models using different levels of supervision. Examples of similar tasks as ours include dehazing and deraining~\cite{chan2023sud, wei2023raindiffusion, ozdenizci2023restoring}, and shadow removal~\cite{guo2023shadowdiffusion}. Conditional diffusion models for underwater image enhancement have also been proposed~\cite{lu2023underwater, tang2023underwater}. Our approach differs from these by focusing on image restoration that inverts the physical model, rather than relying on supervised data that is optimized for image appearance.

\noindent\textbf{Diffusion model as a prior for clean images.}
In addition to conditional generation of images, there is a growing body of research where diffusion models are being used as clean image priors, and image restoration is formulated as posterior sampling~~\cite{choi2021ilvr, chung2022improving, chung2022come, chung2022score, graikos2022diffusion, song2021solving,feng2023score,chung2023diffusion, jalal2021robust, kawar2022denoising, song2023pseudoinverseguided, bansal2023universal}. These include tasks like denoising, inpainting, deblurring, and more general tasks. 
The limited access to clean data in many cases, has led to research on training a prior of clean images, using noisy training data only. In~\cite{aali2023solving, kawar2023gsure, daras2023ambient} a diffusion model prior is trained using noisy data, assuming a known degradation model. Since this setup is not directly applicable in our case, we chose instead to train our prior on clean data that was not taken underwater.

\noindent\textbf{Diffusion model prior for blind image restoration.}
A more challenging task, is to use diffusion models as a clean image prior when the degradation model is unknown or depends on unknown parameters. Several papers have proposed to tackle this problem in different setups. In~\cite{fei2023generative, murata2023gibbsddrm, yang2023pgdiff} the unknown parameters are optimized during the sampling process with a reconstruction objective function. In~\cite{chung2023parallel} a separate prior is trained for the unknown parameters, and sampled in parallel with image sampling. \textbf{Our method differs from the above in that we learn a prior of the main unknown aspect of the degradation model, namely the depth, together with the prior of the variables we want to infer - the image color.} This is done by training a single diffusion model on the joint space of color and depth.

\section{Preliminaries}
\label{sections:Preliminary}

\subsection{Underwater Image Formation}
\label{sections:Preliminary_uw_model}

In water, we observe two wavelength- and distance-dependent effects. First, the \textit{direct} signal reflected from the object is attenuated. 
Second, light is scattered onto the object's line-of-sight (LOS), creating an additive signal termed \textit{backscatter} that increases with distance. The  occluding backscatter layer is independent of the scene content. Thus, the visibility and contrast of further objects is  significantly reduced and their colors are distorted.
 
Following the revised underwater image formation model~\cite{akkaynak2018revised,levy2023seathru},  under ambient illumination image intensity (per pixel, per color channel) is given as:
\begin{equation} \label{eq:uw_model} 
I = J\cdot e^{-\bcD\cdot D}
           + \Binf \cdot \left(1-e^{-\bcB\cdot D}\right)\;\;, 
\end{equation}
where $I$ is the linear image captured by the camera of a scene with range $D$, $J$ is the clear scene that would have been captured had there been no water along the LOS, and $\Binf$ is the  water color at infinity, i.e., the backscatter at areas that contain no objects. The 
two parameters $\bcD$ and $\bcB$ are the attenuation and backscatter coefficients, respectively.

\subsection{Diffusion Models}

Diffusion models have proven to be very effective in capturing the distribution of natural images, given in training data. We describe here the formulation that we use in our work, adopted from~\cite{dhariwal2021diffusion}.
A diffusion model is defined by a Markov chain designed to transform the distribution of real images $x_0$ to a Gaussian distribution $x_T$, by gradually scaling down the image values and adding Gaussian noise, using a schedule determined by the scalar parameters $\alpha_t$,
\begin{equation} \label{eq:diff_forward}
    x_t = \sqrt{\alpha_t} x_{t-1} +  \sqrt{1-\alpha_t}\epsilon, ~~~ 
    \epsilon \sim \mathcal{N}(0, I)\;\;.
\end{equation} 
Based on this \emph{forward} process, an \emph{inverse} process is trained to gradually denoise images starting from a Gaussian distribution, back into the distribution of the original image dataset. This can be formulated as a factorization of the joint distribution over the images in reverse order,  
$p(x_{0:T}) = \prod_{t=1}^T p_t(x_{t-1} \mid x_t)$,
and $p_t(x_{t-1} \mid x_t)$ is approximated by a Gaussian 
\begin{equation} \label{eq:factorization}
p_t(x_{t-1} \mid x_t) \sim \mathcal{N}\big(\mu_\theta(x_t, t), \Sigma_\theta(x_t, t) \big) \;\;,
\end{equation} 
 with parameters $\mu_\theta$ and $\Sigma_\theta$ predicted by a trained neural network conditioned on $x_t$ and $t$. Simulating this process results in samples form the approximated data distribution $p(x_0)$.
A popular implementation first predicts the noise at each time step, and then computes the mean by
\begin{equation} \mu_\theta(x_t,t) =  \frac{1}{\sqrt{\alpha_t}} \left(x_t-\frac{1-\alpha_t }{\sqrt{1-\bar{\alpha}_t}} \epsilon_\theta(x_t,t)\right)\;\;,
\end{equation}
where $\epsilon_\theta(x_t, t)$ is the neural network trained to approximate $\epsilon$, and  $\bar{\alpha}_t = \prod_{s=0}^t \alpha_s$.  The same neural network can also be used to predict a diagonal covariance $\Sigma_\theta(x_t, t)$. 
At any iteration, an estimate of the clean image $x_0$ can be derived by computing the mean of $p(x_0\mid x_t)$:
\begin{equation} \label{eq:clean_x_estimate}
\hat{x}_0(x_t, t) = \frac{1}{\sqrt{\bar{\alpha}_t}} \left(x_t-\sqrt{1-\bar{\alpha}_t} \epsilon_\theta(x_t,t) \right) \;\;.
\end{equation}

The above formulation can also be developed from a \emph{score-based} modeling approach, where it can be shown that $\epsilon_\theta(x_t, t)$ is an approximation to the score function $\nabla_{x_t} \log p_t(x_t)$.
 
\subsection{Posterior Sampling}
\label{sections:Preliminary_post_sampling}
 
Under the score-based view, given an observation $y$ and a likelihood function $p(y | x)$, we can use the same mechanism to sample from the posterior $p(x | y)$, using the posterior score function 
\begin{equation} \label{eq:posterior_score}
    \nabla_{x_t} \log p_t(x_t | y) = \nabla_{x_t} \log p_t(x_t) + \nabla_{x_t} \log p_t(y | x_t) \;\;.
\end{equation}
This idea of adding a conditional signal to the score is also called \emph{guidance}.
The challenge with the second term in Eq.~\ref{eq:posterior_score}, is that usually we are given a likelihood model based on the clean image $x_0$, and not an intermediary noisy image $x_t$. In our case the underwater image formation model transforms a clean image to an underwater observation. Connecting the model to the noisy sample $x_t$ leads to the intractable integral 
\mbox{$p_t(y | x_t) = \int p(y | x_0) p(x_0 | x_t) d x_0$}.

The various works on posterior sampling propose different approximations of this integral.  Some propose to collapse the uncertainty in $y$~\cite{choi2021ilvr, chung2022improving, chung2022come, chung2022score, graikos2022diffusion, song2021solving}. In~\cite{feng2023score} a variational inference approximation is proposed, and in~\cite{chung2023diffusion, jalal2021robust, kawar2022denoising, song2023pseudoinverseguided, bansal2023universal} the likelihood model is computed on either $x_t$ directly, or on the mean \mbox{$\hat{x}_0 = \mathbb{E} [x_0 | x_t] $}, as given in Eq.~\ref{eq:clean_x_estimate}. After some experimentation, we decided to use the latter approximation as formulated in DPS~\cite{chung2023diffusion}:
\begin{equation}\label{eq:app_posterior_score}
\nabla_{x_t} \log p_t(x_t\mid y)  \approx \nabla_{x_t} \log p_t(x_t) +   s \nabla_{x_t} \log p\big(y\mid \hat{x}_0 (x_t, t) \big)\;\;,
\end{equation}
where $s$ is the \emph{guidance scale} used to control the weight of the approximated likelihood term.

\section{Method}
\label{sections:Method}

We use the underwater image formation model (Eq.~\ref{eq:uw_model}) as a forward model that maps the space of natural images $x$ to the space of underwater observed images $y$. We aim to use this forward model to construct a likelihood term $p(y\mid x)$, and use it to sample from the posterior distribution of images (Eq.~\ref{eq:app_posterior_score}). In our case this cannot be implemented directly since the image formation model contains unknown parameters,  the depth $D$ at each pixel, and the water parameters, $\bcD, \bcB, \Binf$, which can differ between scenes. One way to deal with the unknown depth, is to use a monocular depth estimator in a separate first stage, and then use it as part of the forward model. In the results  we discuss this approach and show that it is suboptimal (method variant termed \emph{DA-Osmosis}).

Another approach is to optimize the unknown parameters, including the depth, during sampling. In some recent work, different methods to do this were proposed~\cite{fei2023generative, murata2023gibbsddrm, yang2023pgdiff}. However,  these do not work properly when the unknown parameter is high dimensional, highly complex, and strongly correlated with the target image. In our early experimentation we found that in order to separate the image into a clean image and the effects governed by the depth, we need more than a good image prior. 

Therefore, one of our main contributions, is to train a joint prior on both the color and depth in clean images, and show how it can be used for underwater image restoration.   
Following this we define $x_0 = (J, D)$, where $J$  represents a clear image and consists of the 3 color channels, and $D$ is the depth image.
This allows us to use the following forward model (based on Eq.~\ref{eq:uw_model}) conditioned on the estimated clean image and depth $\hat{x}_0 = (\hat{J}_0, \hat{D}_0)$: 
\begin{equation} \label{eq:forward_model}
f_{\bm{\phi}}(\hat{x}_0) = \hat{J}_0e^{-\bcD \hat{D}_0} + \Binf (1-e^{-\bcB \hat{D}_0})\;\;, 
\end{equation}
where $\bm{\phi} = (\bcD, \bcB, \Binf)$ are the remaining 9 unknown water parameters (a parameter per color channel for each of $\bm{\phi} = (\bcD, \bcB, \Binf)$). We optimize  $\bm{\phi}$ during sampling, in a similar way to GDP~\cite{fei2023generative}, using gradient descent on the image reconstruction loss. The likelihood term used for guidance is defined as a Gaussian around $f_{\bm{\phi}}(\hat{x}_0)$ with a fixed variance.

Training a joint prior on color and depth, not only allows us to use the above forward model, but also exploits the diffusion model capacity to capture the complexity of the depth image and its correlations with the clean color image.  

\subsection{Training the Prior}

\begin{figure}[t]
    \centering
    \includegraphics[width=0.99\linewidth]{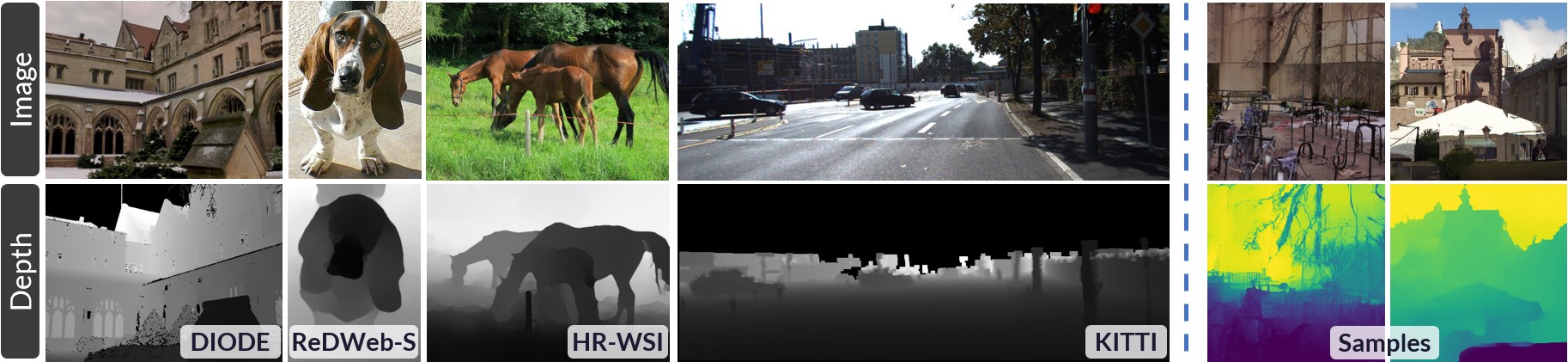}
    \caption{\textbf{[Left]}~Example images from outdoor RGBD datasets used for training our prior. From left to right: DIODE~\cite{diode_dataset}, ReDWeb-S~\cite{liu2021learning}, HR-WSI~\cite{Xian_2020_CVPR}, KITTI~\cite{Geiger2013IJRR}. \textbf{[Right]}~Samples from the trained RGBD prior. The samples demonstrate the inherent correlation between RGB image and depth in our trained RGBD prior.}
    \label{fig:training_data_prior_sample}
\end{figure}

To learn a natural image prior, we train the joint prior model on both color and depth using public RGBD datasets of outdoor scenes that were taken in air. While differing from underwater images, the ability to use large amounts of high quality training data leads to a prior that captures the correlation between color and depth in natural scenes, which as we show, is an important aspect for underwater image restoration. 

For the sake of data efficiency and training time, we start with a pretrained diffusion model trained on ImageNet (we take the unconditional model from~\cite{dhariwal2021diffusion}), and finetune it on RGBD data. We implement this, inspired by DepthGen~\cite{saxena2023monocular}, by replacing the first and last layers of the U-Net to 4 channels rather than 3, and initializing those layers randomly. The datasets we use for finetuning are (see Fig.~\ref{fig:training_data_prior_sample}[left]): DIODE~\cite{diode_dataset}, RedWeb-S~\cite{liu2021learning}, HR-WSI~\cite{Xian_2020_CVPR},    KITTI~\cite{Geiger2013IJRR}, with 16884, 2179, 20378, 23946 pictures, respectively. 

A major challenge in working with RGBD data is to turn the available depth information into a proper image. This includes filling holes and scaling the values to a standard range. Since each of the above datasets were collected in a different manner, we treat each of them differently. This is detailed in the appendix. Fig.~\ref{fig:training_data_prior_sample}[right] shows two samples from the prior that show corresponding color and depth images. While prior samples demonstrate an evident domain gap between in-air and underwater data, our results show that posterior samples are not affected by the gap, and can leverage the strong correlation between depth and color in natural scenes.

\subsection{Sampling from the Posterior}

\begin{figure}[t]
    \centering
    \includegraphics[width=0.98\linewidth]{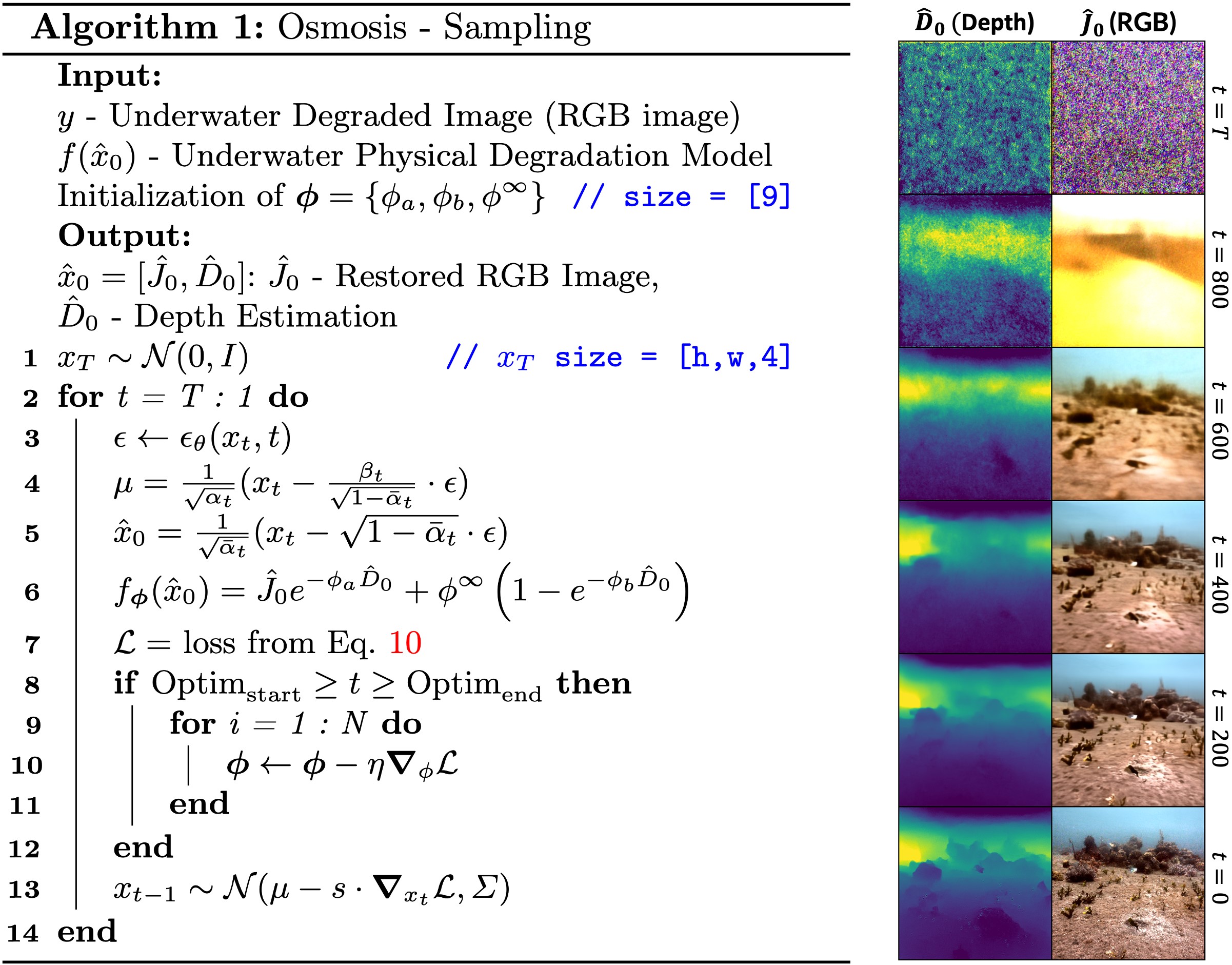}
    \caption{\textbf{Our algorithm.} [Left]~Detailed steps of our algorithm.  [Right]~Example of how $\hat{J}_0, \hat{D}_0$ change during the iterations.} 
    \label{fig:algo_pics}
\end{figure}

Given a trained prior, we perform underwater image restoration by sampling from the posterior with guidance of the image formation model. The sampling process is described in Figs.~\ref{fig:algoprocess},~\ref{fig:algo_pics}.
In each iteration we generate samples of both the image and the depth, while optimizing the water parameters. This results in a gradual update to the estimates of the clean image, and the depth, demonstrated in Fig.~\ref{fig:algo_pics}[right]. 
We adopt the sampling method in DPS~\cite{chung2023diffusion}, using the reconstruction loss, $\big\lVert y-f_\phi(\hat{x}_0) \big\rVert^2_2$, which is the negative log-likelihood formed from the model in Eq.~\ref{eq:forward_model}. Similar to GDP~\cite{fei2023generative}, the remaining unknown parameters $\phi$ are optimized in parallel to the inverse sampling process, using gradient descent on the same loss used for guidance.

Note that one of our main novelties is that the RGBD prior is used as an inherent part of the iterative method and not as a stand-alone depth estimator. The depth is estimated together with the image in \emph{every} iteration, guided by the underwater model, see, e.g.,  Fig.~\ref{fig:algo_pics}[right]. As demonstrated in the results, this improves the restoration quality over using a fixed estimated depth of the scene.

Running posterior sampling, guided by a reconstruction loss only, we observe two problems. First, for pixels with large depth values, the reconstructed image is dominated by the backscatter, making the estimation of the clean pixel color values unstable. Second, the color values can shift outside the valid range, causing color saturation.  To overcome the first issue we multiply the reconstruction loss in every pixel by the estimated depth value (without passing the gradients of this operation, denoted by \emph{sg}):
\begin{equation}\label{eq:rec_loss}
\mathcal{L}_{\rm rec} = \big\lVert sg(\hat{D}_0) \cdot \left(y-f_\phi(\hat{x}_0)\right) \big\rVert^2_2 \qquad .
\end{equation}
In order to overcome the second issue, we introduce two auxiliary losses. The first penalizes RGB values outside the valid range $[-1, 1]$, and the second encourages the average values of each channel to approach the middle of the color range (in line with the \emph{gray world assumption}). We implement the two auxiliary losses as:
\begin{eqnarray*} \label{eq:aux_loss}
\mathcal{L}_{\rm val} = \lambda_v \sum_{i,c} \max \left(\left|\hat{J}_0(i, c) \right| - T_v , 0 \right)^2,  ~~~
\mathcal{L}_{\rm avrg} = \lambda_a \sum_c \left| \sum_i \hat{J}_0(i, c) - T_a \right| 
\end{eqnarray*}
where $\hat{J}_0(i, c)$ is the value of color channel $c$ of pixel $i$, assuming the valid color range is $[-1, 1]$, $T_v$ and $T_a$ are thresholds set to values close to 1 and 0 respectively, and $\lambda_v, \lambda_a$ are the scalar weights of both losses. The total loss  is then:
\begin{equation}\label{eq:loss}
\mathcal{L} = \mathcal{L}_{\rm rec} + \mathcal{L}_{\rm val} + \mathcal{L}_{\rm avrg}\;\;.
\end{equation}

At each iteration we compute the gradient of the loss with respect to $x_t$, which forms the log-likelihood gradient in the posterior score (Eq.~\ref{eq:app_posterior_score}), and the gradient with respect to $\phi$ which is used to update the parameters in the underwater model.  
In order to stabilize the optimization, we apply gradient clipping by value, to the gradients of $x_t$, and we run the optimization of $\phi$ only in some of the sampling steps (defined by the values ${\rm Optim}_{\rm start}$ and ${\rm Optim}_{\rm end}$), running $N$ iterations of gradient descent in each step. 
When the gradient of $x_t$ is applied in the sampling step, it is multiplied by a guidance scale $s$. We find that using a smaller scale for the depth channel leads to better results. This makes sense as while all channels are treated the same in the prior, the depth has a different role in the forward model (specifically it is used inside an exponent), and this can lead to a different gradient scale.
More details on the implementation are given in the appendix.

\section{Results}

We use a prior on $256 \times 256$ images, with sampling time of  about 3 minutes per image on an Nvidia A100 GPU. We present here selected results and analysis. Please refer to the appendix for a complete set.

\subsubsection{Real-World Scenes.}
We present an extensive comparison on real-world linear images from the datasets SQUID~\cite{berman2020underwater}, SeaThru~\cite{akkaynak2019sea}, SeaThruNerf~\cite{levy2023seathru}, and additional images acquired in different locations in the world, the Indian and Pacific Oceans, and Mediterranean, and Red Seas. Images are white-balanced as pre-processing.  All runs use the same set of parameters. Fig.~\ref{fig:teaser} shows several results of both image restoration and depth estimation. 

\begin{figure*}
    \centering
    \includegraphics[width=0.99\linewidth]{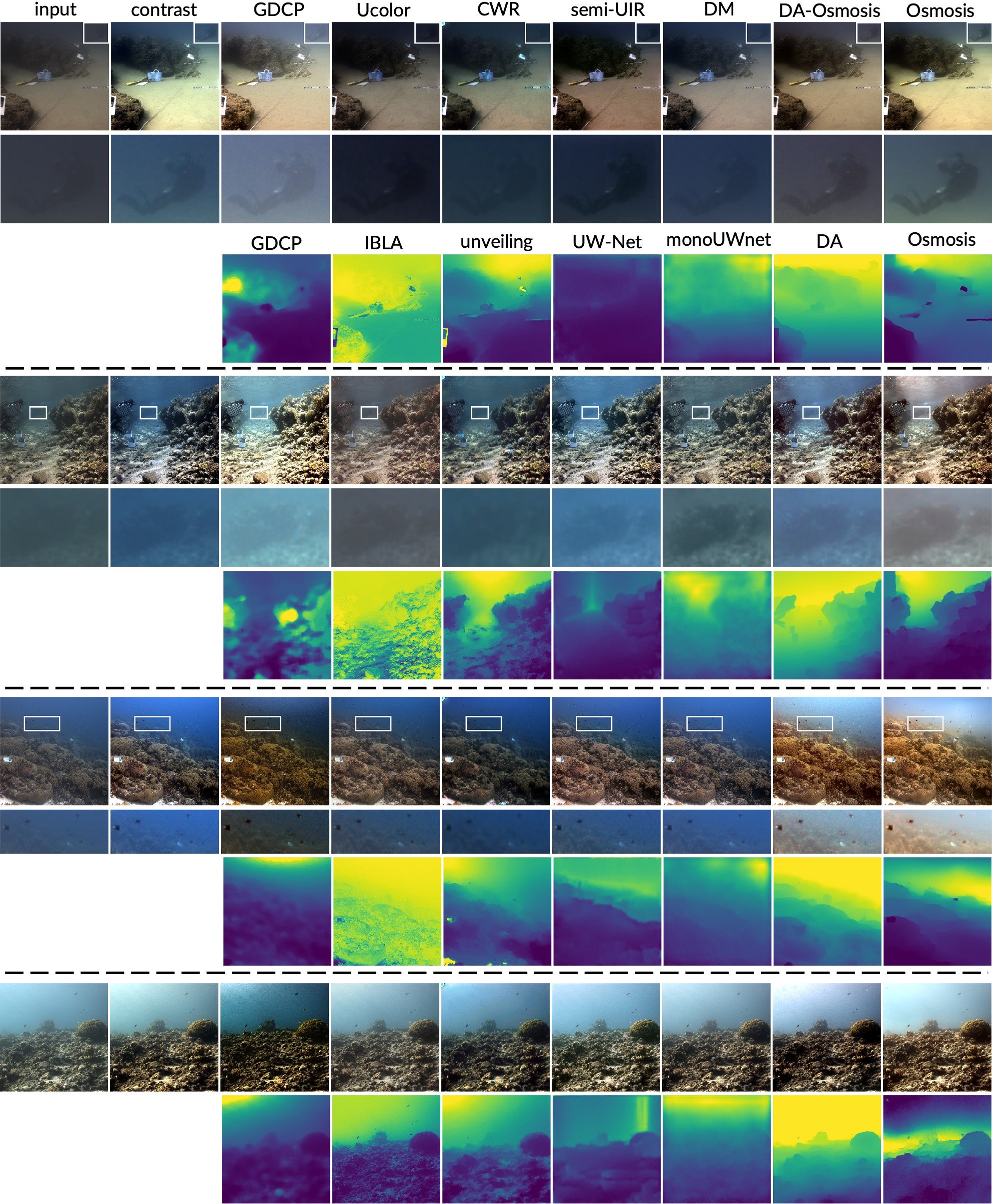}
    \caption{\textbf{Real-world restoration results.} From left to right: white-balanced input, contrast stretch, GDCP~\cite{peng2018generalization}, Ucolor~\cite{li2021underwater}, CWR~\cite{han2022underwater}, semi-UIR~\cite{huang2023contrastive}, DM~\cite{tang2023underwater}, Depth Anything~\cite{yang2024depth} - Osmosis, Osmosis (ours). Zoom-in colored rectangles emphasize far objects that have higher contrast in our results. \textbf{Real-world depth results.} From left to right: GDCP~\cite{peng2018generalization}, IBLA~\cite{peng2017underwater}, unveiling~\cite{bekerman2020unveiling}, UW-Net~\cite{gupta2019unsupervised}, monoUWnet~\cite{amitai2023self}, Depth Anything~\cite{yang2024depth}, Osmosis (ours). Our depth results are smoother and less affected by object gradients. \textbf{The reader is encouraged to zoom-in.}}
    \label{fig:rw_color}
\end{figure*}

We compare \textbf{image restoration} with the following methods: CWR~\cite{han2022underwater}, DM~\cite{tang2023underwater}, FUnIE-GAN~\cite{islam2020fast}, GDCP~\cite{peng2018generalization}, IBLA~\cite{peng2017underwater}, MMLE~\cite{zhang2022underwater}, semi-UIR~\cite{huang2023contrastive}, Ucolor~\cite{li2021underwater}, USUIR~\cite{fu2022unsupervised}, UW-Net~\cite{gupta2019unsupervised}, waternet~\cite{li2019underwater}, and USe-ReDI-Net~\cite{varghese2023self}. For compactness, Fig.~\ref{fig:rw_color} summarizes the results for a chosen subset of methods.  The complete comparison is shown in the appendix. Among previous methods, we found GDCP to be the most consistent. Our result recovers the full range of the scenes, specifically improving contrast of objects that are further away (note the zoomed-in objects in the far areas, e.g., the diver in row 1). Our method recovers vibrant and consistent colors also in further areas. For example, in rows 2,3,4 note that the background rocks and sand appear bluish in all methods except of ours.

We also compare to a method that restores the RGB using a fixed depth that is pre-estimated using an external SOTA method, \emph{depth anything} (DA)~\cite{yang2024depth}. We use the pre-estimated depth in our method's pipeline instead of our gradually estimated depth. We term this method \emph{DA-Osmosis}, and the results show that our method restores the far areas with better color, emphasizing the advantage of using our joint RGBD prior. 

We compare \textbf{depth estimation}  with GDCP~\cite{peng2018generalization}, IBLA~\cite{peng2017underwater}, unveiling~\cite{bekerman2020unveiling}, UW-Net~\cite{gupta2019unsupervised}, and monoUWnet~\cite{amitai2023self}, and depth anything (DA)~\cite{yang2024depth}. In GDCP~\cite{peng2018generalization} and unveiling~\cite{bekerman2020unveiling} we compute the depth from the output transmission maps.  Our depth estimations have more details and better explain the scenes (Fig.~\ref{fig:rw_color}). 

In addition, we conducted a non-reference quantitative comparison using the \textbf{MUSIQ}~\cite{Ke_2021_ICCV} measure on $50$ real-world images presented in the paper and appendix. Our method achieved the highest score. The quantitative results are summarized in Table~\ref{tab:musiq_table_50_images}.

\begin{figure}[t]
    \centering
    \includegraphics[width=0.99\linewidth]{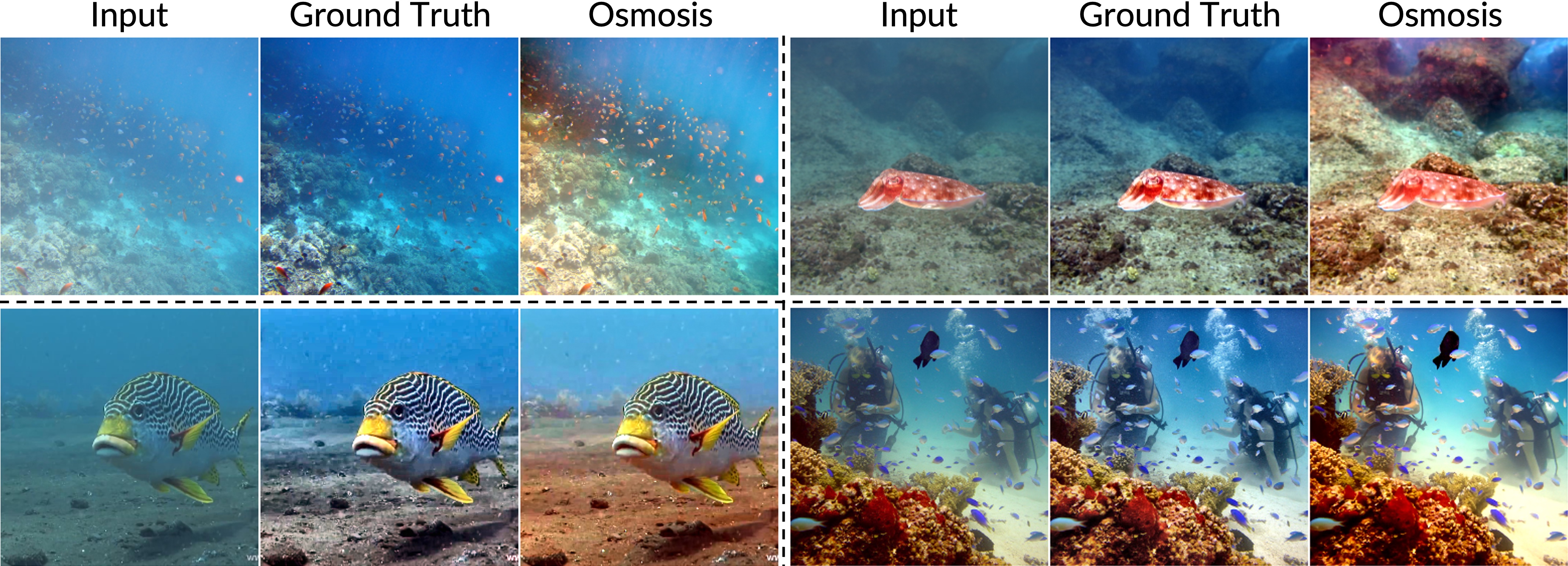}
    \caption{\textbf{Example results on UIEB~\cite{li2019underwater},} a dataset comprised from non-linear images with ground truth produced by different enhancement algorithms.
    Our method yields \textbf{better} results than the dataset's ground truth. See for example the color of the sand, the orange color of the fish in the top-left, and the divers' authentic skin color in the bottom-right.}
    \label{fig:uieb}
\end{figure}

The UIEB dataset~\cite{li2019underwater} is sometimes used for quantitative evaluation of image restoration. However, it is collected from various sources and is not linear. Thus it is not suitable for physics-based methods like ours. 
In addition, the  `ground truth' of the UIEB dataset is chosen from results of previous methods and it is not the real ground truth. 
Fig.~\ref{fig:uieb} shows four examples (out of many) where our result removes more water effects than the ground truth, even despite the non-linearity of the input.

\subsubsection{Simulation.}

We conducted a simulation following~\cite{huang2023contrastive,li2020underwater} using all the 1449 images from the NYU-v2 RGBD dataset~\cite{silberman2012indoor} with randomly varying water parameters and generated the underwater images with the image formation model in Eq.~\ref{eq:uw_model}. For fairness in comparison with other methods we applied $\bcD=\bcB$.

The quantitative results are summarized in Table~\ref{tab:simulation_results}. Our method substantially outperforms other methods in the image restoration metrics of PSNR, SSIM, and LPIPS. 
We emphasize that our prior was \textbf{not} trained on this data, or indoor data at all.  Moreover, two of the  methods~\cite{huang2023contrastive,li2020underwater} we outperform were trained on this data. 

\begin{table}[ht]
    \centering
    \footnotesize{
    \begin{minipage}[t]{0.45\linewidth}
        \raggedright
        \centering
        \begin{tabular}{c c} \hline
            Method &  MUSIQ$\uparrow$ \\ \hline
            input &  51.59 \\ 
            contrast stretch &  53.75 \\ 
            CWR~\cite{han2022underwater} &  39.63 \\ 
            DM~\cite{tang2023underwater} &  54.64 \\ 
            FUnIE-GAN~\cite{islam2020fast} &  42.70 \\ 
            GDCP~\cite{peng2018generalization} &  56.25 \\ 
            IBLA~\cite{peng2017underwater} &  54.42 \\ 
            MMLE~\cite{zhang2022underwater} &  \underline{56.26} \\ 
            semi-UIR~\cite{huang2023contrastive} &  54.99 \\ 
            Ucolor~\cite{li2021underwater} &  47.73 \\ 
            USUIR~\cite{fu2022unsupervised} &  52.44 \\ 
            UW-Net~\cite{gupta2019unsupervised} &  46.02 \\ 
            waternet~\cite{li2019underwater} &  53.64 \\ 
            unveiling~\cite{bekerman2020unveiling} &  50.83 \\ 
            DA-osmosis~\cite{yang2024depth} &  51.23 \\ \hline
            \textbf{osmosis (ours)} &  \textbf{56.62} \\ \hline
        \end{tabular}
        \caption{Non-reference quantitative comparison on $50$ real-world images using \textbf{MUSIQ}~\cite{Ke_2021_ICCV}. Our method achieves the highest score.}
        \label{tab:musiq_table_50_images}
    \end{minipage}%
    \hfill
    \begin{minipage}[t]{0.45\linewidth}
        \raggedright
        \centering
        \begin{tabular}{c c c c}
            \hline
             Method &  PSNR$\uparrow$ & SSIM$\uparrow$ & LPIPS$\downarrow$ \\
            \hline
            contrast stretch                  & 17.13 &  0.83 &  0.11 \\ 
            CWR~\cite{han2022underwater}      & 16.93 &  0.79 &   0.20 \\ 
            DM~\cite{tang2023underwater}     & 17.41 &  0.82 &   0.12 \\ 
            FUnIE-GAN~\cite{islam2020fast}          & 17.64 &  0.77 &   0.21 \\ 
            GDCP~\cite{peng2018generalization} & 12.41 &  0.71 & 0.16  \\  
            IBLA~\cite{peng2017underwater}     & 15.07 &  0.70 &  0.19 \\  
            MMLE~\cite{zhang2022underwater}    & 17.00 &  0.74 &   0.17 \\ 
            semi-UIR~\cite{huang2023contrastive}   & 17.82 &  0.83 &   0.12 \\ 
            Ucolor~\cite{li2021underwater}       & 17.92 &  0.83 &   0.10 \\ 
            USUIR~\cite{fu2022unsupervised}     & 16.76 &  0.80 &   0.18 \\ 
            UW-Net~\cite{gupta2019unsupervised}  & 18.04 &  0.75 &  0.26 \\ 
            waternet~\cite{li2019underwater}       & 17.27 &  0.82 &  0.11 \\ 
            unveiling~\cite{bekerman2020unveiling} & 16.34 &  0.79 &   0.18  \\ 
            \hline
            ablation \#1                         & 21.09 &  0.86 &  0.09  \\  
            ablation \#2                           & \underline{22.17} & \underline{0.88} &  0.07  \\ 
            ablation \#3                           & 22.00 & \underline{0.88} &  \textbf{0.06}  \\ \hline 
            \textbf{osmosis (ours)}                        & \textbf{22.74} & \textbf{0.89} &  \textbf{0.06} \\ \hline
        \end{tabular}
        \caption{Quantitative comparison on the simulation. Our method achieves best scores in image restoration.}
        \label{tab:simulation_results}
    \end{minipage}
    }
\end{table}

\subsubsection{Ablation.}

\begin{figure*}[t]
    \centering
    \includegraphics[width=0.99\linewidth]{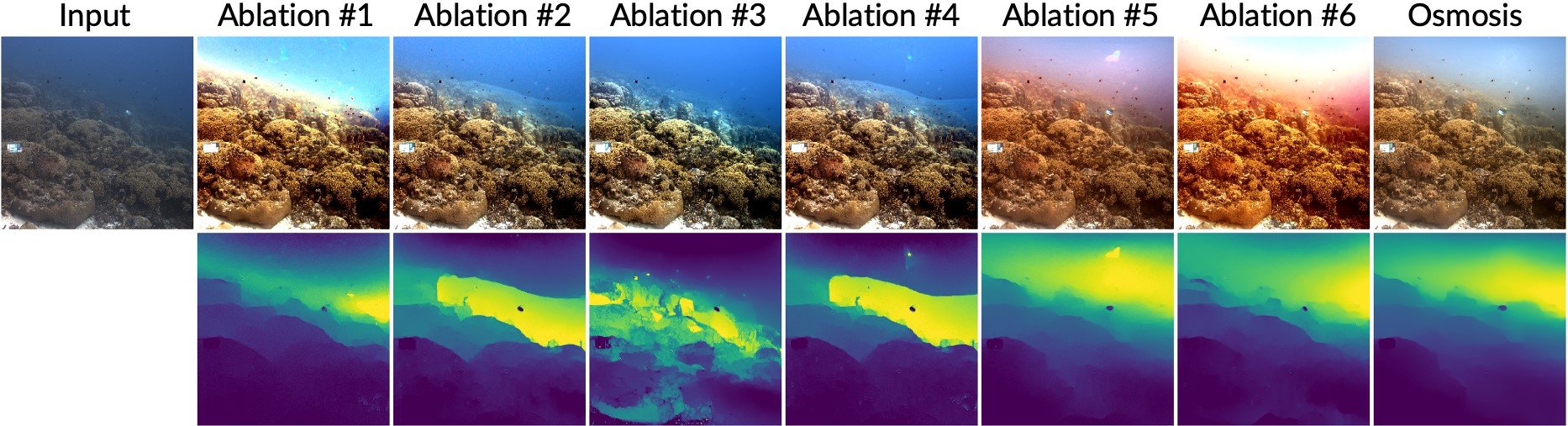}
    \caption{\textbf{Ablation.} Ablations \#1,\#5,\#6 show the importance of the losses we added to the reconstruction loss. Without $\mathcal{L}_{\rm val}$ (\#6), the colors tend to ``explode'' and oversaturate. Without $\mathcal{L}_{\rm avrg}$ (\#5) the colors sometimes skew towards pink/purple. In  \#2, the further areas are not restored well, because the loss is not weighted by the depth $D$. In \#3 the guidance scale is the same for the RGB and depth channels. This harms the depth reconstruction. In  \#4 we set $\bcD=\bcB$ (separately per color channel). Since this is an inaccurate model, the restoration in further areas is harmed.}
    \label{fig:ablation}
\end{figure*}

To demonstrate the effect of different components in our methods, we conduct an ablation study of the following variants: 
\noindent\textbf{1.}  $\mathcal{L}=\mathcal{L}_{\rm rec}$ (instead of Eq.~\ref{eq:loss});
\noindent\textbf{2.}   Removing weighting by $\hat{D}_0$ in Eq.~\ref{eq:rec_loss}
\noindent\textbf{3.}   Same guidance scale $s$ for all channels; 
\noindent\textbf{4.}  $\bcD=\bcB$;
\noindent\textbf{5.}  
$\mathcal{L} = \mathcal{L}_{\rm rec} + \mathcal{L}_{\rm val}$ (remove $\mathcal{L}_{\rm avrg}$ from Eq.~\ref{eq:loss});
\noindent\textbf{6.}  $\mathcal{L} = \mathcal{L}_{\rm rec} + \mathcal{L}_{\rm avrg} $ (remove $\mathcal{L}_{\rm val}$  from Eq.~\ref{eq:loss}).
 Numerical results on the simulation for variants \#1-\#3 are presented in Table~\ref{tab:simulation_results}. 
 Ablations \#4-\#6 are shown only on real world images since in the simulation we do not use $\mathcal{L}_{\rm avrg}$, and $\bcD=\bcB$. Fig.~\ref{fig:ablation} presents the results of all variants on one of the scenes. We see that the additional losses are important to prevent color saturation and shift. Increasing the loss weight with depth improves restoration in further areas. Separating guidance scales between the RGB and depth channels improves depth reconstruction. Setting $\bcD\neq\bcB$ extends the range of the restoration.

\section{Discussion}

In this work we demonstrated how to harness the strength of a new RGBD diffusion prior to achieve state-of-the-art results on underwater image restoration. To do this we solved several challenges: i)~because of lack of clean underwater image data we use datasets of scenes in air; ii)~we notice that the color prior does not suffice to guide restoration, therefore, we add the scene depth to the prior;
iii)~We use the physical image formation model to guide restoration, and also estimate the water parameters in the process. 
This results in the most comprehensive single underwater image restoration method to-date. It does not train on any underwater images and therefore does not overfit to any. It gives higher weight to further objects and therefore is superior in reconstructing all the details of the scene. Jointly solving for the depth results in excellent depth estimation from monocular images.
 
There is a domain gap between the prior (trained on in-air data) and underwater data. This is intentional. The beauty of our method is that by incorporating the underwater model in the sampling it succeeds even \emph{without training} on underwater data (that is very hard to obtain). Thus, it is not specialized for a certain type of water or objects, and it learns the color distribution of natural images not degraded by water.

Like all diffusion models with U-Nets, our method is limited by a fixed resolution and long running time. This could potentially be improved using other network architectures and sampling methods.

\noindent\small\textbf{Acknowledgements.}
The research was funded by Israel Science Foundation grant $\#1951/23$, Israeli Ministry of Science and Technology grants $\#1001577600$ \& $\#1001593851$,  EU Horizon 2020 research and innovation programme GA $101094924$ (ANERIS), the  Leona M. and Harry B. Helmsley Charitable Trust, and the Maurice Hatter Foundation.
We thank Dr.~Derya~Akkaynak and Dr.~Matan~Yuval for
substantial data contribution, Amir Dayan for the 
paper's name and Meirav Keidar for graphics design.


\ifarxiv 
\clearpage 
\appendix
\section*{Appendix}

\label{sec:appendix}
In this appendix we give more details about the implementation- prior training and optimization, and provide multiple additional results. All figures are placed at the end of the document.

\section{Data Preprocessing and Training}
Here we describe the different datasets used for training, and the preprocessing performed on each.

\textbf{KITTI}~\cite{Geiger2013IJRR} - 23946 images. Depth information is from Lidar measurement and is sparse. We interpolate it into dense depth images using~\cite{kitti_dense_map}. We then normalize by the maximum measurement value of $80$ meters. When computing the loss we mask out depth pixels of remaining holes and non-depth information like the sky.

\textbf{DIODE}~\cite{diode_dataset} - 16884 images. Depth information is from a high quality laser scanner. We normalize by the maximum value of the depth sensor which is 350. Valid depth masks are supplied, and used when computing the loss.

\textbf{HR-WSI}~\cite{Xian_2020_CVPR}- 20378 images. Computed with stereo cameras. The data is a relative disparity with values between 0 and 1, but without an absolute normalization value. We compute the depth as $1 - disparity$. Valid mask are provided and used when computing the loss.

\textbf{Red-Web-S}~\cite{liu2021learning} - 2179 images - The depth information is computed from a model's prediction and is already dense and normalized. 

We crop and resize the images to get a size of 256x256. In KITTI we crop-out the upper part of the image which contain only sky to get a 256 height, and then use different random horizontal crops of 256. In all other datasets we resize the smaller dimension into 256, and crop the other dimension at the center.  We perform additional data augmentations by horizontal and vertical flips.

\textbf{Details on the training process}.  We train both $\epsilon_{\theta}(x_{t},~t)$ and $\Sigma_{\theta}(x_{t},~t)$ (defined in Sec.~3.2 in the paper). We use the two losses suggested in~\cite{dhariwal2021diffusion}, $L_{\rm simple}$ is MSE with a mask for the non-valid pixels (e.g., holes and horizon) in the image, and $L_{\rm vlb}$, a Variational Lower Bound.

\section{Implementation Details}
We give a list of implementation details and specific values used in the experiments. These values are used throughout all experiments except for some different values in the simulation (stated in bold below). 
\begin{enumerate}
\item Both in training and in sampling we use 1000 sampling steps between 0 to $T=1$. 
\item We used a linear schedule for the diffusion noise variance, in the following range: 
\begin{equation*}
\alpha_t = 1-\beta_t , ~~ \beta_{0}=1e^{-4}  , ~\beta_{1}=2e^{-2}
\end{equation*}
\item We use a U-Net architecture which was suggested in~\cite{dhariwal2021diffusion}, for 256x256 input. In order to handle RGBD data, we made modifications for the first and last convolution layers. The first convolution layer gets 4 channels as input instead of 3 channels, and the last convolution output is 4 channels instead of 3.  These two layers are initialized at random before finetuning.
\item Before the depth map is used in the underwater model, it is linearly scaled from the range $[-1, 1]$ into the range $[0.56, 3.36]$ using the function:
\begin{equation*}
g(\hat{D}_0)=1.4\cdot(\hat{D}_0+1.4)
\end{equation*}
In the \textbf{simulation}, this is mapped to a $[0, 1]$ range instead, using 

\mbox{$g(\hat{D}_0)=0.5\cdot(\hat{D}_0+1)$}.
The model's depth range is tuned as a standard hyperparameter and is the same for all the real-world images.
\item In the reconstruction loss, the weight is computed according to the same linearly scaled depth using a `stop-gradient' operator.
\begin{equation*}
g(sg(\hat{D}_0))=1.4\cdot(sg(\hat{D}_0)+1.4)
\end{equation*}
This is used both for real world and the simulation experiments.
\item For all real world experiments we use a guidance scale, separated to each of the RGBD channels, as following:
red: 7, green: 7, blue: 7, depth: 0.9. In the \textbf{simulation} the values are: red: 4, green: 4, blue: 4, depth: 1.
\item The weights of the auxiliary losses are:
\begin{equation*}
    \lambda_v=20, ~~~ \lambda_a=0.5
\end{equation*}
In the \textbf{simulation} we do not use $\mathcal{L}_{\rm avrg}:$ 
\begin{equation*}
    \lambda_v=40, ~~~ \lambda_a=0
\end{equation*}
\item The thresholds used in the auxiliary losses are:~~\mbox{$T_v = 0.7, ~~~ T_a = 0.5$}
\item The threshold of gradient clipping we used is $0.005$. In the \textbf{simulation} The threshold of gradient clipping we used is $0.001$.
\item We initialized the water parameters to:
\begin{enumerate}
    \item $\phi_a: 1.1,0.95,0.95$
    \item $\phi_b : 0.95, 0.8, 0.8$
    \item  $\phi^{\infty}: 0.14, 0.29, 0.49$
\end{enumerate}
In the \textbf{simulation} we use the simpler model where \mbox{$\phi_a=\phi_b$}, and initialize according to $\phi_a$ above, and  $\phi^{\infty}$ is initialized to $[0.2,0.4,0.7]$.
\item The optimization schedule of $\phi$ was set to run from step $t=0.7$ down to step $t=0$, with 20 gradient descent iterations at each step. 
\begin{equation*}
    {\rm Optim}_{\rm start}: 0.7, ~~{\rm Optim}_{\rm end}: 0,  ~~N: 20
\end{equation*}
\end{enumerate}

\section{Real World Results}

In this appendix we present a total of \textbf{48} results of Osmosis on challenging real world scenes. 
For 16 real-world scenes (8 of them were presented in the mai   n paper) we provide extensive comparisons with other methods in Figs.~\ref{fig:full_results_fig1},~\ref{fig:full_results_fig5},~\ref{fig:full_results_fig_else1},~\ref{fig:full_results_fig_else2}. Additionally, we provide 16 additional scenes of real-world restored RGB and depth maps generated by Osmosis in Fig.~\ref{fig:seathru_squid_results}. We also include results on additional 16 images in Fig.~\ref{fig:sup_consistency}. 

\subsection{Full comparisons}
In Figs.~\ref{fig:full_results_fig1},~\ref{fig:full_results_fig5},~\ref{fig:full_results_fig_else1},~\ref{fig:full_results_fig_else2} we present results of the complete suite of comparison methods on the all the real-world scenes presented in the main paper and on additional scenes:  a)~Input, b)~contrast stretch, c)~GDCP~\cite{peng2018generalization},  d)~IBLA~\cite{peng2017underwater}, e)~unveiling~\cite{bekerman2020unveiling}, f)~UW-Net~\cite{gupta2019unsupervised}, g)~waternet~\cite{li2019underwater}, h)~Ucolor~\cite{li2021underwater}, i)~MMLE~\cite{zhang2022underwater}, j)~CWR~\cite{han2022underwater},  k)~FUnIE-GAN~\cite{islam2020fast},  l)~USUIR~\cite{fu2022unsupervised},  m)~semi-UIR~\cite{huang2023contrastive},     n)~DM~\cite{tang2023underwater} , o)~DA-Osmosis~\cite{yang2024depth} , p)~Osmosis~(ours). For USe-ReDI-Net~\cite{varghese2023self} there is no released code, therefore comparison is shown in Fig.~\ref{fig:sup_iccv23} on the 3 linear scenes that were presented in~\cite{varghese2023self}.

\subsection{More results}
We show results on additional scenes from SQUID~\cite{berman2020underwater} and Seathru~\cite{akkaynak2019sea} in Fig.~\ref{fig:seathru_squid_results}. 

\section{Extended Ablations}

In addition to the example provided in Fig.~\ref{fig:ablation} of the main paper, three additional examples are presented here for the same ablation study. 
To demonstrate the effect of all parts of our methods, we conduct an ablation study of the following variants: \\
\noindent\textbf{1.}  $\mathcal{L}=\mathcal{L}_{\rm rec}$ (instead of Eq.~\ref{eq:loss}). \\
\noindent\textbf{2.}   Removing weighing by $\hat{D}_0$ in Eq.\ref{eq:rec_loss}. \\
\noindent\textbf{3.}   Same guidance scale $s$ for all channels. \\
\noindent\textbf{4.}  $\bcD=\bcB$  \\
\noindent\textbf{5.}  
$\mathcal{L} = \mathcal{L}_{\rm rec} + \mathcal{L}_{\rm val}$ (remove $\mathcal{L}_{\rm avrg}$ from Eq.~\ref{eq:loss}).
  \\
\noindent\textbf{6.}  $\mathcal{L} = \mathcal{L}_{\rm rec} + \mathcal{L}_{\rm avrg} $ (remove $\mathcal{L}_{\rm val}$  from Eq.~\ref{eq:loss}).  \\

Numerical results on the simulation for variants \#1-\#3 are  presented in Table~1 in the main paper. Ablations \#4-\#6 are shown only on real world images since in the simulation we do not use $\mathcal{L}_{\rm avrg}$, and $\bcD=\bcB$. Fig.~\ref{fig:sup_ablation} presents the results of all variants on 3 scenes presented in Fig.~\ref{fig:teaser} in the main paper. We see that the additional losses are important to prevent color saturation and shift. Increasing the loss weight with depth improves restoration in further areas. Separating guidance scales between the RGB and depth channels improves depth reconstruction. Setting $\bcD\neq\bcB$ extends the range of the restoration.

\section{Simulation}

Fig.~\ref{fig:sup_simulation} depicts a method comparison on several scenes from the simulation. We can see that our method cleans the entire range of the image, while previous methods recover mostly the nearby areas.

\section{Consistency Analysis}

As diffusion is a random process, we test our method's consistency in two different experiments. 
Fig.~\ref{fig:sup_consistency} demonstrates a consistency experiment, where we ran our method on several images of the same scene from several viewpoints. We can see the resulting scene has similar appearance, and the depth is consistent.

Fig.~\ref{fig:sup_seed_consistency} shows multiple results on the same image using a different random seed each time. We see that our results are very similar even with different seeds, showing the strength of our formulation and losses.

\section{Results on Haze}

Fig.~\ref{fig:haze} shows our results on several haze images. Although our method was not designed and optimized for haze, the haze  is removed in the restored images, the scenes have vivid colors and the depth maps are good. 
To run our method on these images we performed a ``degamma'' operation on them ($I^{2.2}$), and set $\bcD=\bcB$ constant for all color channels (one parameter instead of 6).

\section{Non-linear images}

Every physics-based method expects as input linear images. We demonstrate the effect of inputting non-linear images in Fig.~\ref{fig:supp_non_linear}. We took the uncompressed non-linear images of the linear images used as input in Fig.~\ref{fig:teaser} in the main paper, performed white-balance on them, and used them as input for our method. We can see in the results that the range of the restoration is smaller (i.e., the restoration stops at some point), for both color and depth. In addition, the colors are skewed.

\section{Failure Cases}

Fig.~\ref{fig:sup_failure_rw} demonstrates 3 failure cases on real-world linear images. In example~1, there is an artifact in the top-right corner. In example~2 the restored colors are reddish and saturated. In example~3 there is a pinkish hue, especially in the horizon (``sky'') area.

We have noticed that, sometimes, the sky is not recognized as the most far area of the restored depth map (e.g., example~2 in Fig.~\ref{fig:sup_failure_rw}). A possible reason is that during training, in some of the data, the sky is masked out and replaced by a value of 0. Although these pixels are masked out in the loss as well, they can still affect the prior through the input images. The effect can also have been amplified by the vertical flipping in our data augmentation, resulting in many cases where the top of the image has smaller depth.

\section{Negative Results and Abandoned Directions}
We give here a list of directions that were tried and either gave worse results to the ultimate method we use, or did not show any promise in early experimentation and was therefore abandoned. The goal of this section is to share more information in order to give a bigger picture of our experimentation process. Many of the results here were not thoroughly examined and therefore should be treated as such. 

\begin{enumerate}
    \item We tried computing the gradient w.r.t  $x_0$ rather than $x_t$ for the likelihood score. This is similar to the Backward Universal Guidance suggested in~\cite{bansal2023universal}. This resulted in noisy images. 
    \item We ran a few initial experiments using Dynamic Guidance Scheme~\cite{yang2023pgdiff,voynov2023sketch}, and changing the guidance scale during the sampling process as suggested in~\cite{bansal2023universal, graikos2022diffusion}. We did not see significant improvement, but we believe further experimentation in this direction could be fruitful. 
    \item We tried to completely stop the guidance for the last few iterations as suggested in~\cite{voynov2023sketch}, because we noticed that in some cases the restored colors in early stages of the sampling looked better than in the end. This resulted in improvements to the colors, but the geometric details were less preserved.
    \item We tried an inner optimization of $\hat{x}_0$ for several iterations at each as suggested in~\cite{bansal2023universal}. This resulted in the generated image being too noisy, which can perhaps be explained as the sampling process going too far off the manifold of the prior.
    \item We tried running several iterations of sampling and $\phi$ optimization for the same time step~\cite{murata2023gibbsddrm},  or per-step self-recurrence~\cite{bansal2023universal}. The aim of this feature is to keep the restored image committed to the prior and in addition strengthening the guidance. In our case, this resulted in somewhat distorted colors and in a worse estimation of the depth geometry.
    \item We considered different annealing scheduling of the sampling time~\cite{graikos2022diffusion}. We did not perform vast experimentation with this. In our results the color restoration was more conservative, i.e. more stable but fixing less of the water effects.
    \item We tried to clip the $\hat{x}_0$ image values in the forward model, instead of clipping the gradients. This lead to color saturation or de-saturation (colors getting closer to black). We found that gradient clipping in addition to the $\mathcal{L}_{\rm val}$ auxiliary loss gave better stability to the color restoration.
    
\end{enumerate}

\begin{figure*}[t]
    \centering
    \includegraphics[width=1\linewidth]{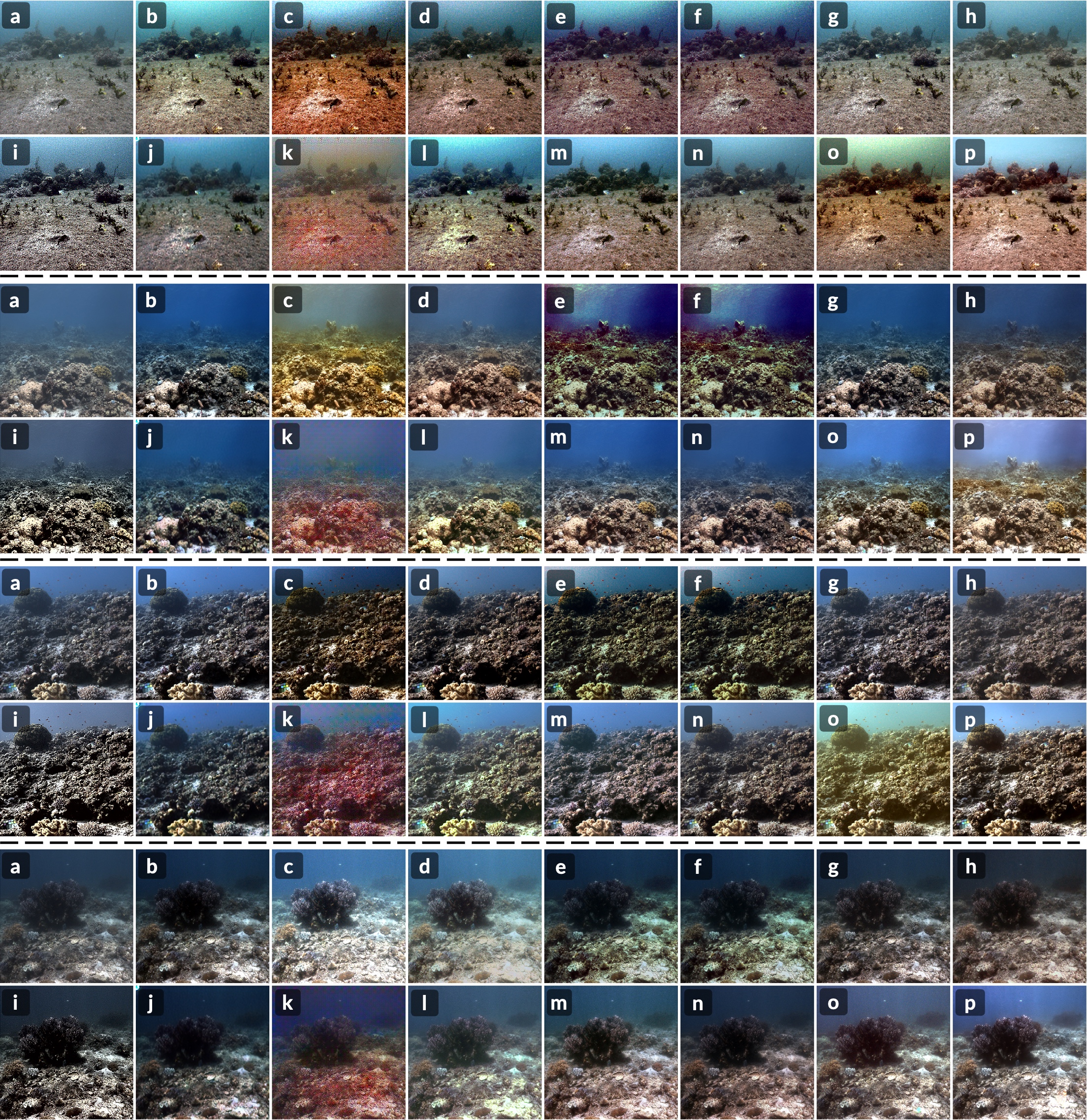}
    \caption{\textbf{Comparisons with all the methods on the scenes presented in Fig.~\ref{fig:teaser} in the main paper.} a)~Input, b)~contrast stretch, c)~GDCP~\cite{peng2018generalization},  
    d)~IBLA~\cite{peng2017underwater}, e)~unveiling~\cite{bekerman2020unveiling}, f)~UW-Net~\cite{gupta2019unsupervised}, g)~waternet~\cite{li2019underwater}, h)~Ucolor~\cite{li2021underwater}, i)~MMLE~\cite{zhang2022underwater}, j)~CWR~\cite{han2022underwater},  k)~FUnIE-GAN~\cite{islam2020fast},  l)~USUIR~\cite{fu2022unsupervised},  m)~semi-UIR~\cite{huang2023contrastive},     n)~DM~\cite{tang2023underwater} , o)~DA-Osmosis~\cite{yang2024depth} , \textbf{p)~Osmosis (ours)}. Our restorations have the best colors and recovery range. \textbf{The reader is encouraged to zoom-in.}}
    \label{fig:full_results_fig1}
\end{figure*}


\begin{figure*}[t]
    \centering
    \includegraphics[width=1\linewidth]{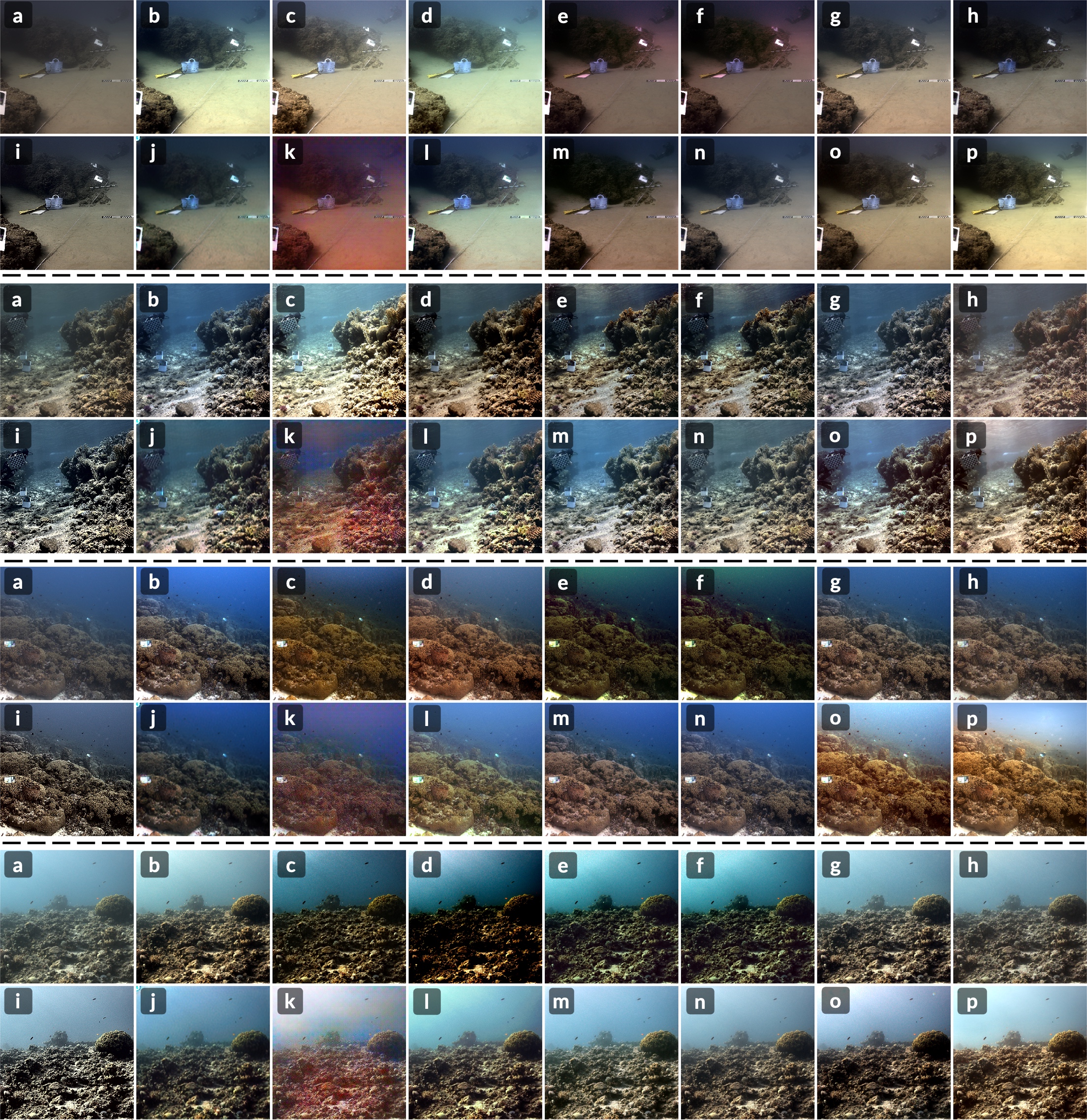}
    \vspace{-0.2cm}
    \caption{\textbf{Comparisons with all the methods on the scenes presented in Fig.~\ref{fig:rw_color} in the main paper.} a)~Input, b)~contrast stretch, c)~GDCP~\cite{peng2018generalization},  
    d)~IBLA~\cite{peng2017underwater}, e)~unveiling~\cite{bekerman2020unveiling}, f)~UW-Net~\cite{gupta2019unsupervised}, g)~waternet~\cite{li2019underwater}, h)~Ucolor~\cite{li2021underwater}, i)~MMLE~\cite{zhang2022underwater}, j)~CWR~\cite{han2022underwater},  k)~FUnIE-GAN~\cite{islam2020fast},  l)~USUIR~\cite{fu2022unsupervised},  m)~semi-UIR~\cite{huang2023contrastive},     n)~DM~\cite{tang2023underwater} , o)~DA-Osmosis~\cite{yang2024depth} , \textbf{p)~Osmosis (ours)}. Our restorations have the best colors and recovery range. \textbf{The reader is encouraged to zoom-in.} }
    \label{fig:full_results_fig5}
\end{figure*}

\begin{figure*}[t]
    \centering
    \includegraphics[width=1\linewidth]{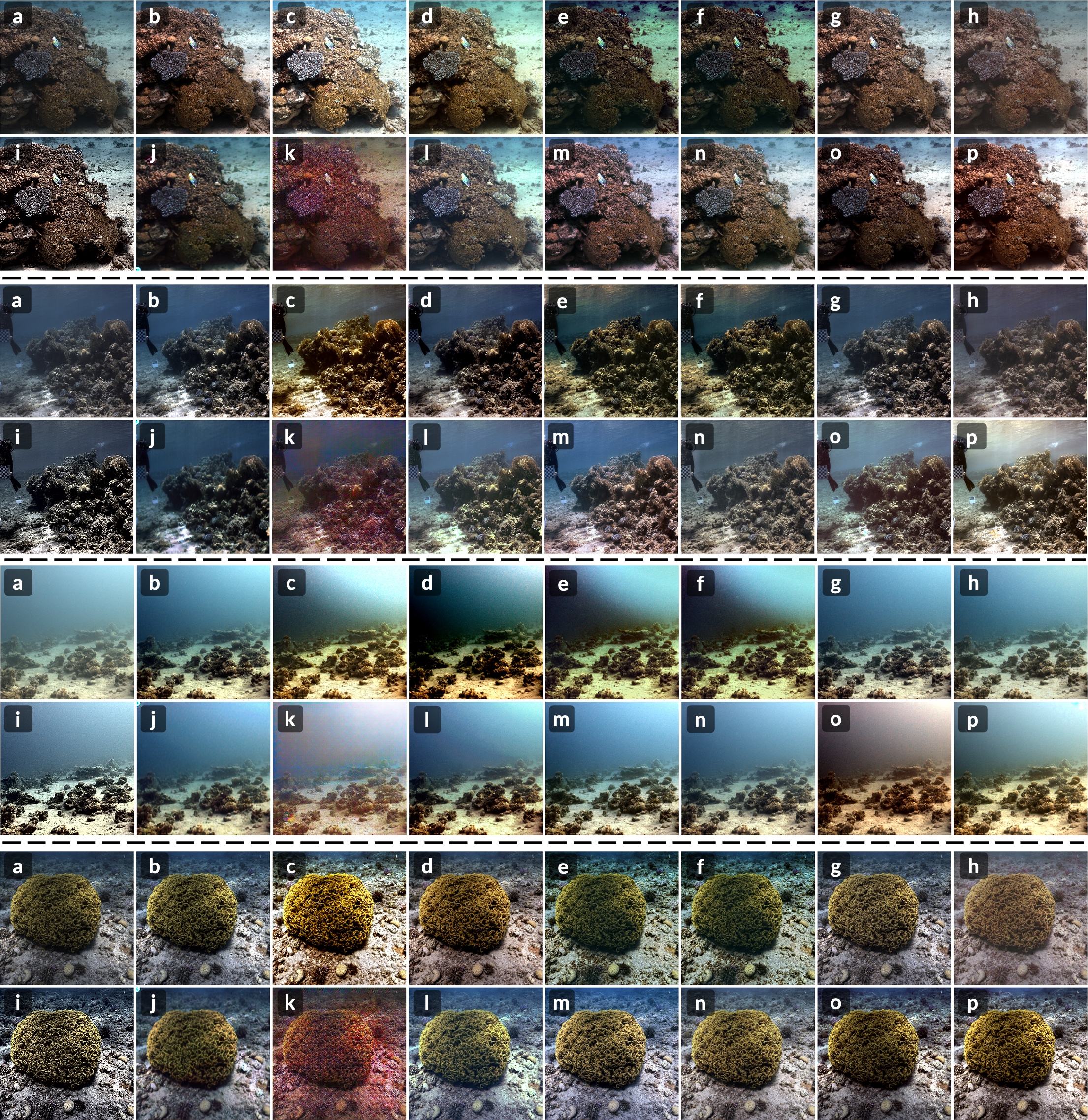}
    \vspace{-0.2cm}
    \caption{\textbf{Comparisons with all the methods on additional scenes.} a)~Input, b)~contrast stretch, c)~GDCP~\cite{peng2018generalization},  
    d)~IBLA~\cite{peng2017underwater}, e)~unveiling~\cite{bekerman2020unveiling}, f)~UW-Net~\cite{gupta2019unsupervised}, g)~waternet~\cite{li2019underwater}, h)~Ucolor~\cite{li2021underwater}, i)~MMLE~\cite{zhang2022underwater}, j)~CWR~\cite{han2022underwater},  k)~FUnIE-GAN~\cite{islam2020fast},  l)~USUIR~\cite{fu2022unsupervised},  m)~semi-UIR~\cite{huang2023contrastive},     n)~DM~\cite{tang2023underwater} , o)~DA-Osmosis~\cite{yang2024depth} , \textbf{p)~Osmosis (ours)}. Our restorations have the best colors and recovery range. \textbf{The reader is encouraged to zoom-in.} }
    \label{fig:full_results_fig_else1}
\end{figure*}

\begin{figure*}[t]
    \centering
    \includegraphics[width=1\linewidth]{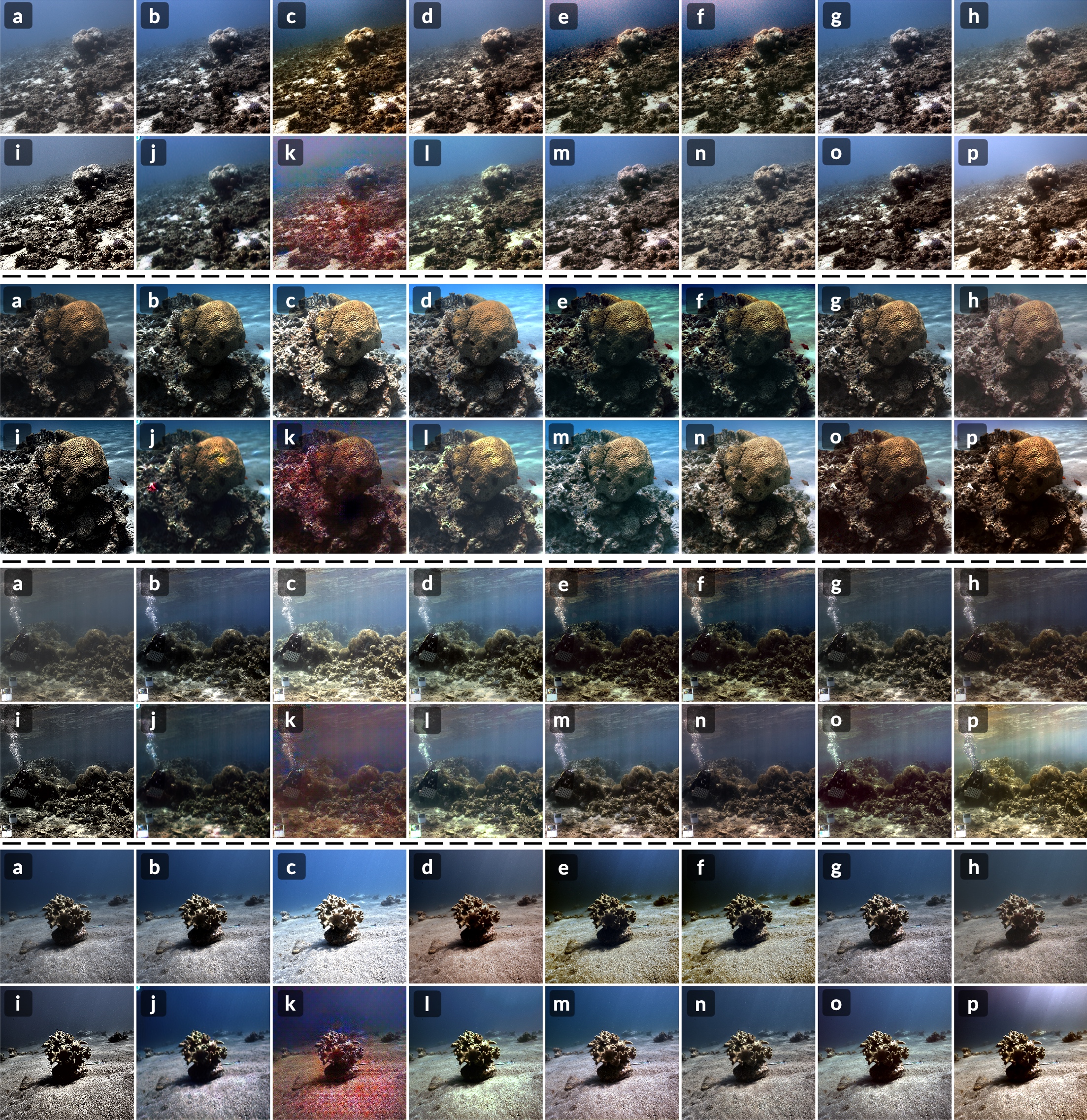}
    \vspace{-0.2cm}
    \caption{\textbf{Comparisons with all the methods on additional scenes.} a)~Input, b)~contrast stretch, c)~GDCP~\cite{peng2018generalization},  
    d)~IBLA~\cite{peng2017underwater}, e)~unveiling~\cite{bekerman2020unveiling}, f)~UW-Net~\cite{gupta2019unsupervised}, g)~waternet~\cite{li2019underwater}, h)~Ucolor~\cite{li2021underwater}, i)~MMLE~\cite{zhang2022underwater}, j)~CWR~\cite{han2022underwater},  k)~FUnIE-GAN~\cite{islam2020fast},  l)~USUIR~\cite{fu2022unsupervised},  m)~semi-UIR~\cite{huang2023contrastive},     n)~DM~\cite{tang2023underwater} , o)~DA-Osmosis~\cite{yang2024depth} , \textbf{p)~Osmosis (ours)}. Our restorations have the best colors and recovery range. \textbf{The reader is encouraged to zoom-in.} }
    \label{fig:full_results_fig_else2}
\end{figure*}

\begin{figure}[t]
    \centering
    \includegraphics[width=0.6\linewidth]{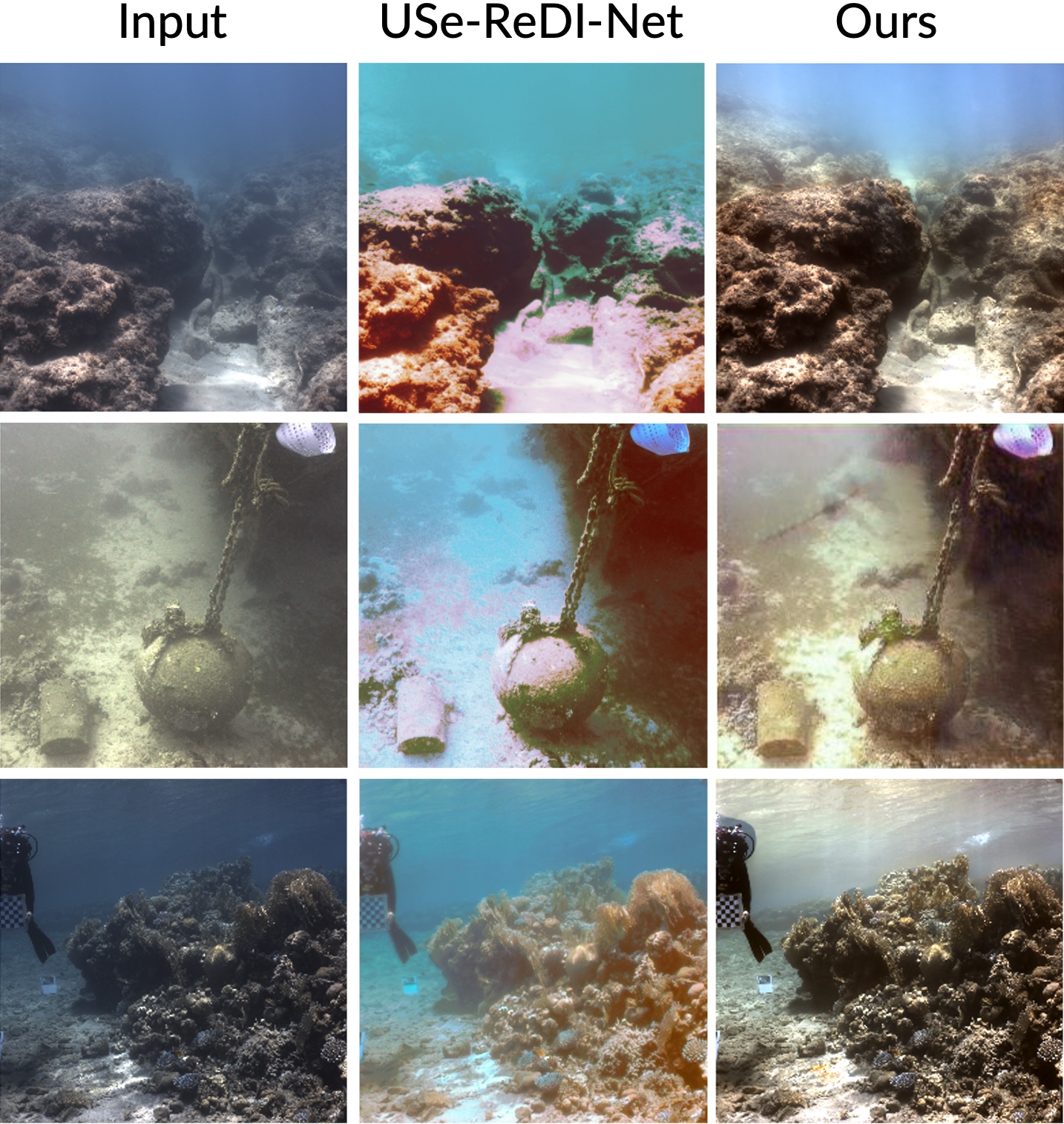}
    \caption{\textbf{Comparisons with~\cite{varghese2023self}.} Since there is no published code for~\cite{varghese2023self} we can only compare with results published in the paper on linear scenes. Our restorations have consistent colors across the scenes.}
    \label{fig:sup_iccv23}
\end{figure}

\begin{figure*}[t]
    \centering
    \includegraphics[width=0.95\linewidth]{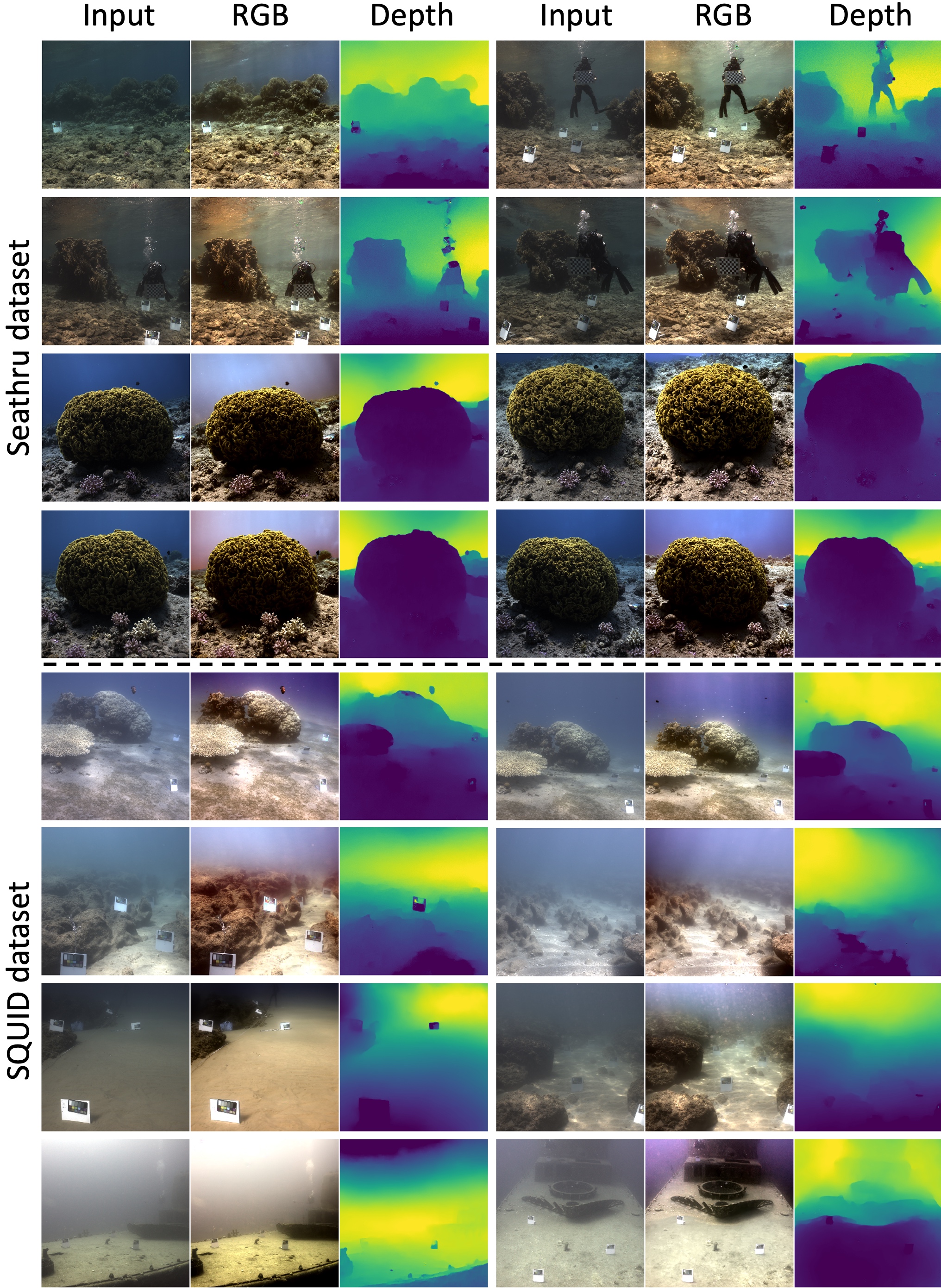}
    \vspace{-0.2cm}
    \caption{\textbf{Additional Results.} Osmosis results on several scenes from Seathru~\cite{akkaynak2019sea} and SQUID~\cite{berman2020underwater} datasets.
    The restored image displays significantly less water effects, while maintaining consistent colors. 
    We note that the estimated depth for very bright objects (e.g. the color boards) tends to be too close. This could be a result of the irregularity of having such objects within natural scenes, considering the training data. In any case this does not have a noticeable effect on the restored image. \textbf{The reader is encouraged to zoom-in.} }
    \label{fig:seathru_squid_results}
\end{figure*}

\begin{figure*}[t]
    \centering
    \includegraphics[width=1\linewidth]{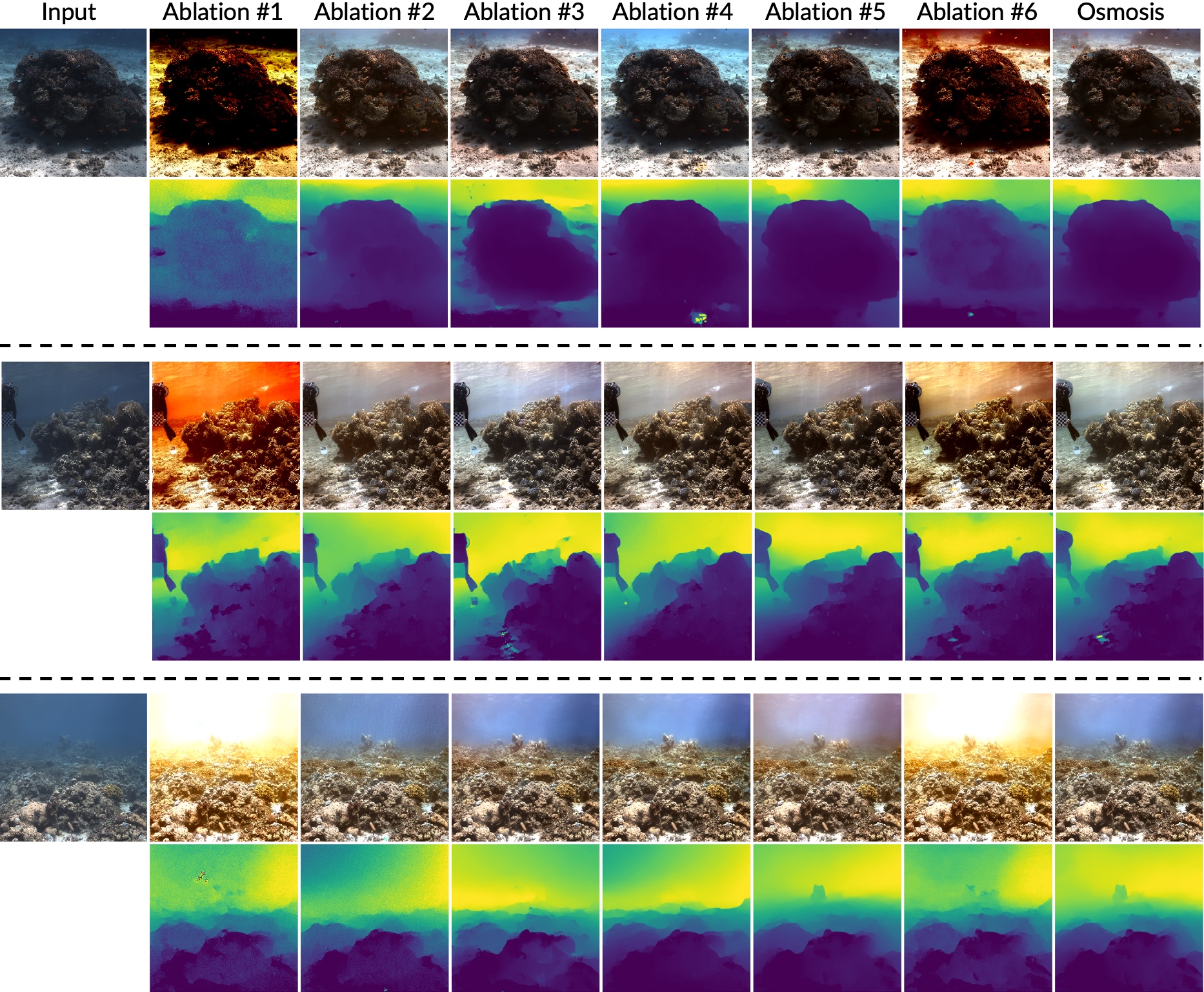}
    \caption{\textbf{Ablation results on several real-world images.} Ablations \#1,\#5,\#6 demonstrate the importance of the losses we added to the reconstruction loss. Without $\mathcal{L}_{\rm val}$ (\#6), the colors tend to ''explode`` and oversaturate. Without $\mathcal{L}_{\rm avrg}$ (\#5) the colors sometimes skew towards pink/purple. In ablation \#2, we can see that the further areas are not restored well, this is because the loss is not weighed by the depth $D$. In ablation \#3 the guidance scale is the same for the RGB and depth channels. This harms the depth reconstruction. In ablation \#4 we set $\bcD=\bcB$ (separately per color channel). Since this is an inaccurate model, the restoration in further areas is harmed.}
    \label{fig:sup_ablation}
\end{figure*}

\begin{figure*}[t]
    \centering
    \includegraphics[width=0.95\linewidth]{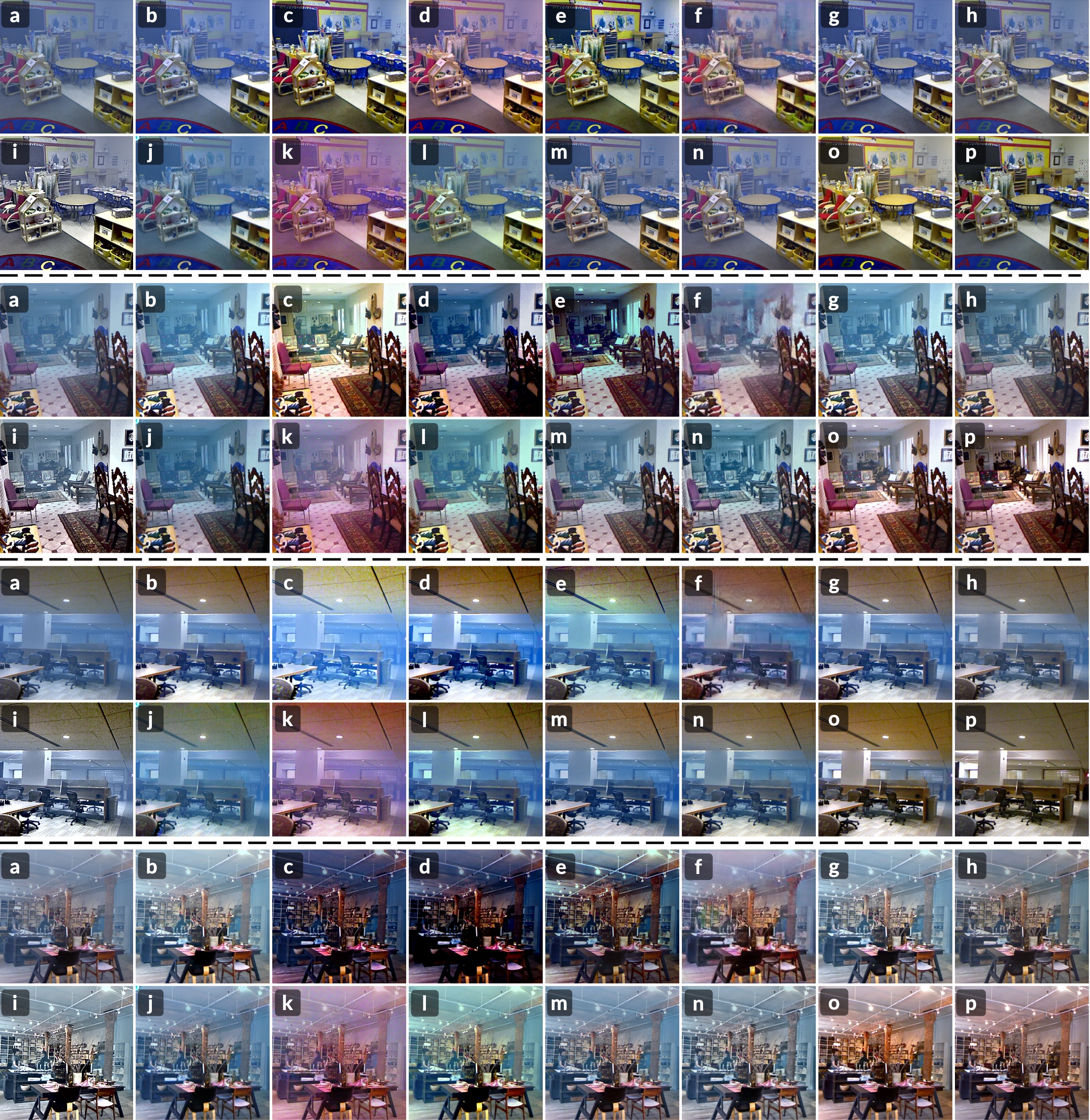}
    \caption{\textbf{Several color restoration results on simulated images.} a)~Input, b)~contrast stretch, c)~GDCP~\cite{peng2018generalization},  
    d)~IBLA~\cite{peng2017underwater}, e)~unveiling~\cite{bekerman2020unveiling}, f)~UW-Net~\cite{gupta2019unsupervised}, g)~waternet~\cite{li2019underwater}, h)~Ucolor~\cite{li2021underwater}, i)~MMLE~\cite{zhang2022underwater}, j)~CWR~\cite{han2022underwater},  k)~FUnIE-GAN~\cite{islam2020fast},  l)~USUIR~\cite{fu2022unsupervised},  m)~semi-UIR~\cite{huang2023contrastive},     n)~DM~\cite{tang2023underwater} ,  \textbf{o)~Osmosis (ours)}, p)~Ground-truth. Our color restoration achieves highest PSNR, and cleans more areas in the scenes. }
    \label{fig:sup_simulation}
\end{figure*}

\begin{figure*}[t]
    \centering
    \includegraphics[width=1\linewidth]{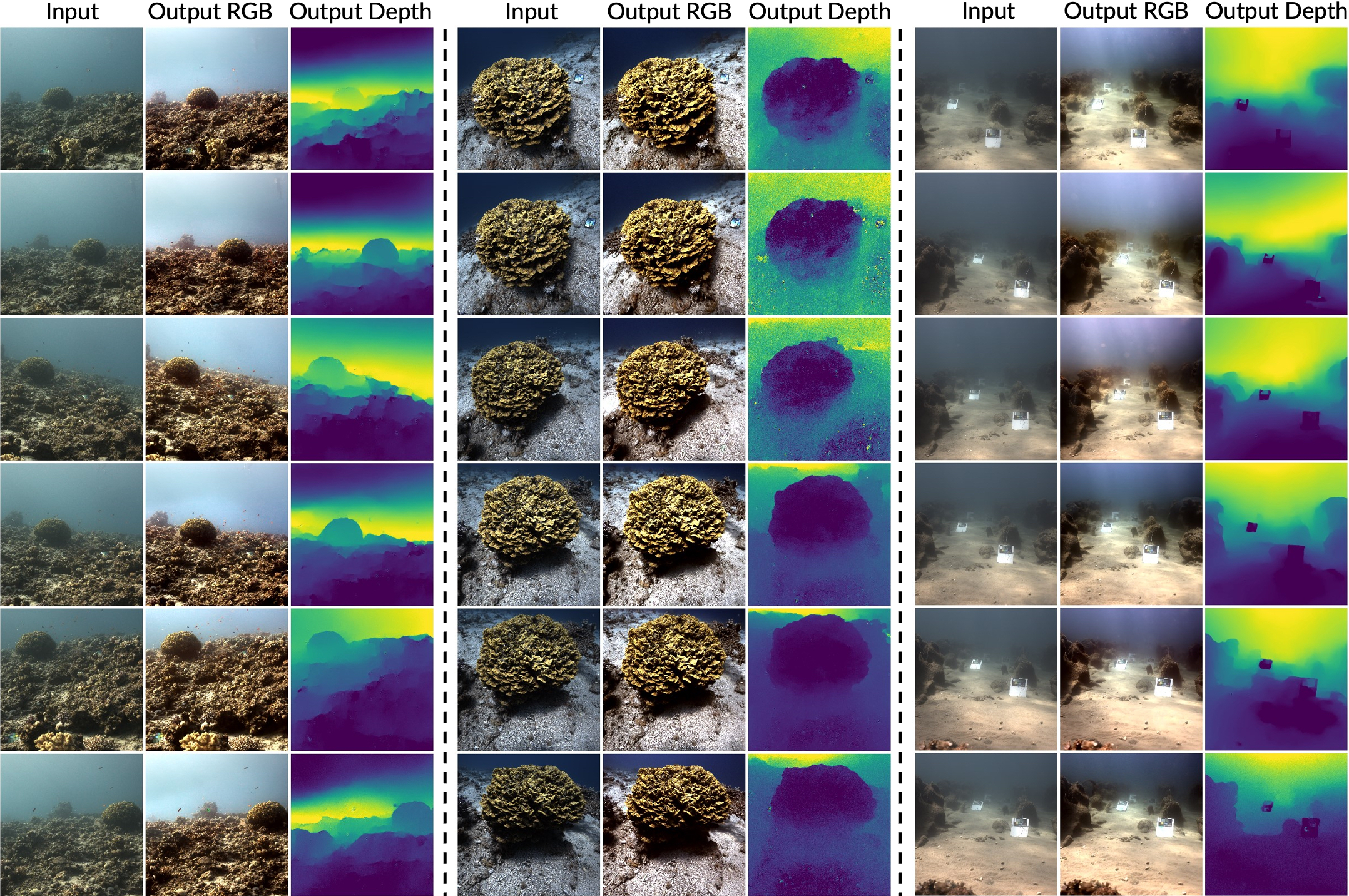}
    \caption{\textbf{Consistency of results.} We ran our method on several images of the same scene from several viewpoints. The resulting restorations have similar appearances in terms of color, and the depth is consistent.}
    \label{fig:sup_consistency}
\end{figure*}

\begin{figure*}[t]
    \centering
    \includegraphics[width=1\linewidth]{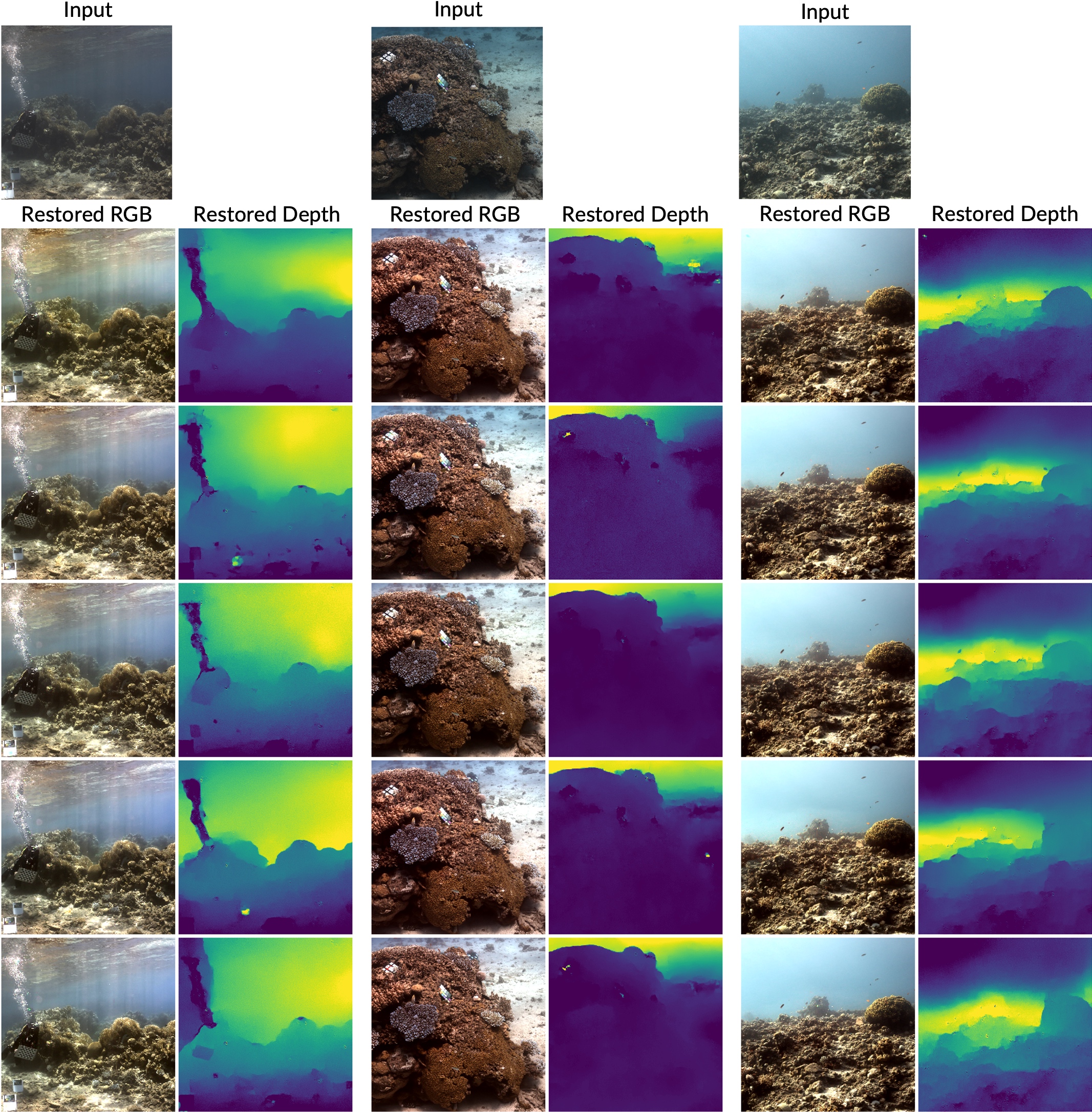}
    \caption{\textbf{Consistency of results given different random seeds for 3 different scenes.} Each row presents results with different random seeds. We see that the results are very similar in both color and depth, showing the strength of our method's optimization procedure.}
    \label{fig:sup_seed_consistency}
\end{figure*}

\begin{figure}[t]
    \centering
    \includegraphics[width=0.7\linewidth]{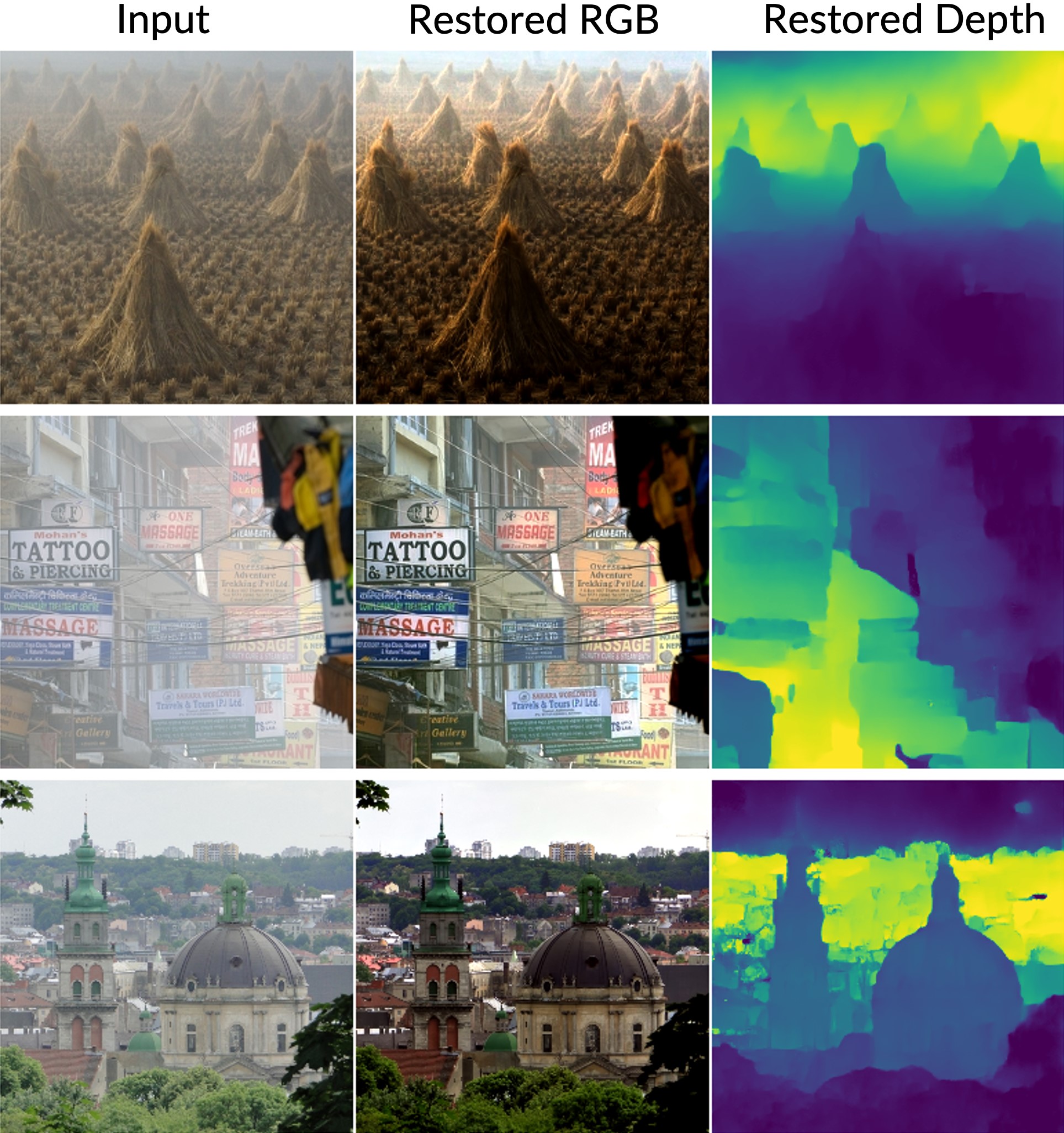}
    \caption{\textbf{Results on haze.} Though not our main goal, we demonstrate that our method can potentially work also on haze images, when setting $\bcD=\bcB$ (identical for all color channels) in Eq.~\ref{eq:forward_model} in the main paper. Note the vivid colors and the depth maps.}
    \label{fig:haze}
\end{figure}

\begin{figure*}[t]
    \centering
    \includegraphics[width=1\linewidth]{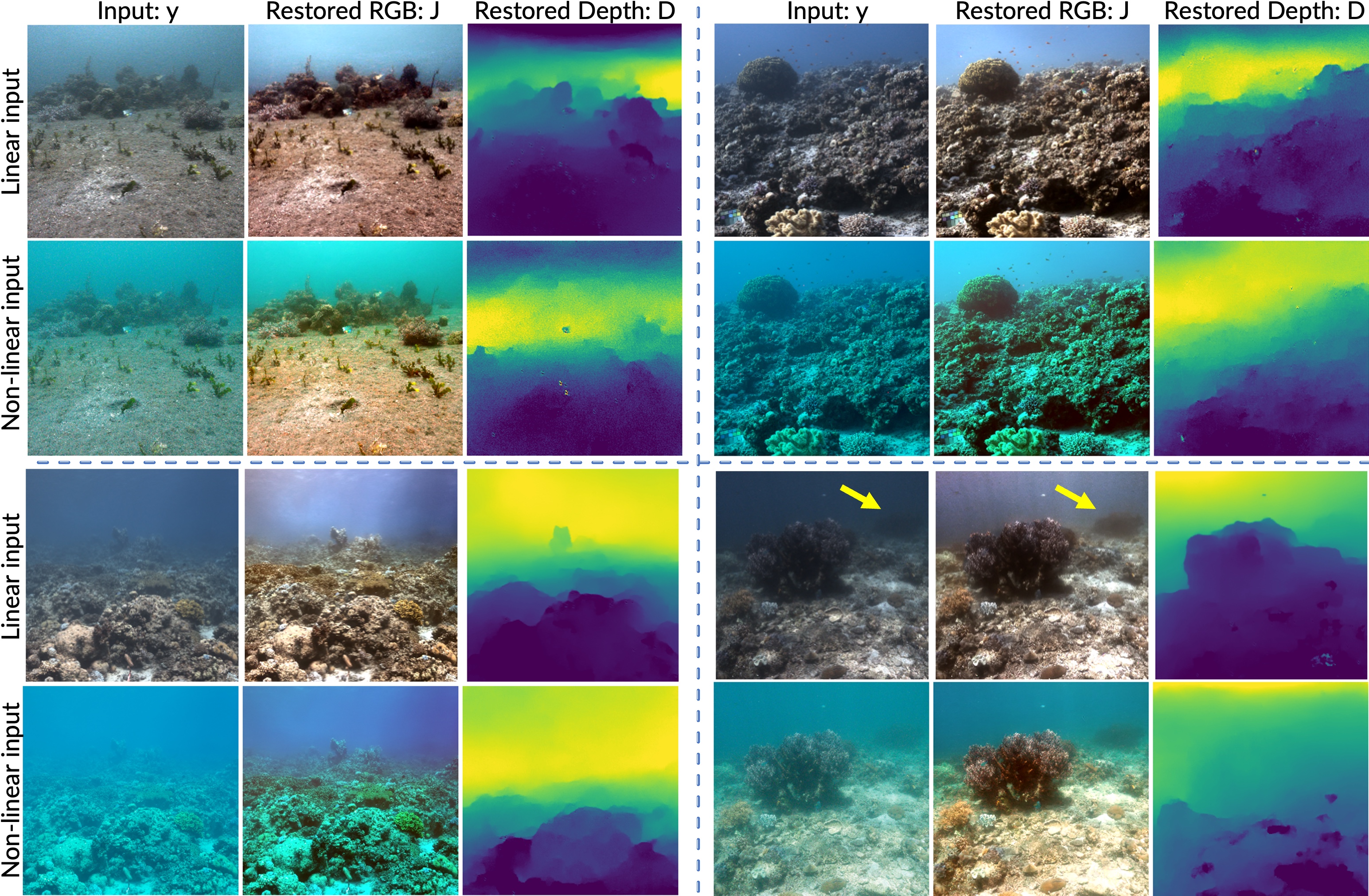}
    \caption{\textbf{Effect of non-linear input.}  We took the non-linear camera jpgs for the scenes presented in Fig.~1 in the main paper and ran our method after white-balancing. We see that the reconstruction range of both the colors and the depth maps is limited, and the restored colors are skewed.}
    \label{fig:supp_non_linear}
\end{figure*}

\begin{figure}[t]
    \centering
    \includegraphics[width=0.6\linewidth]{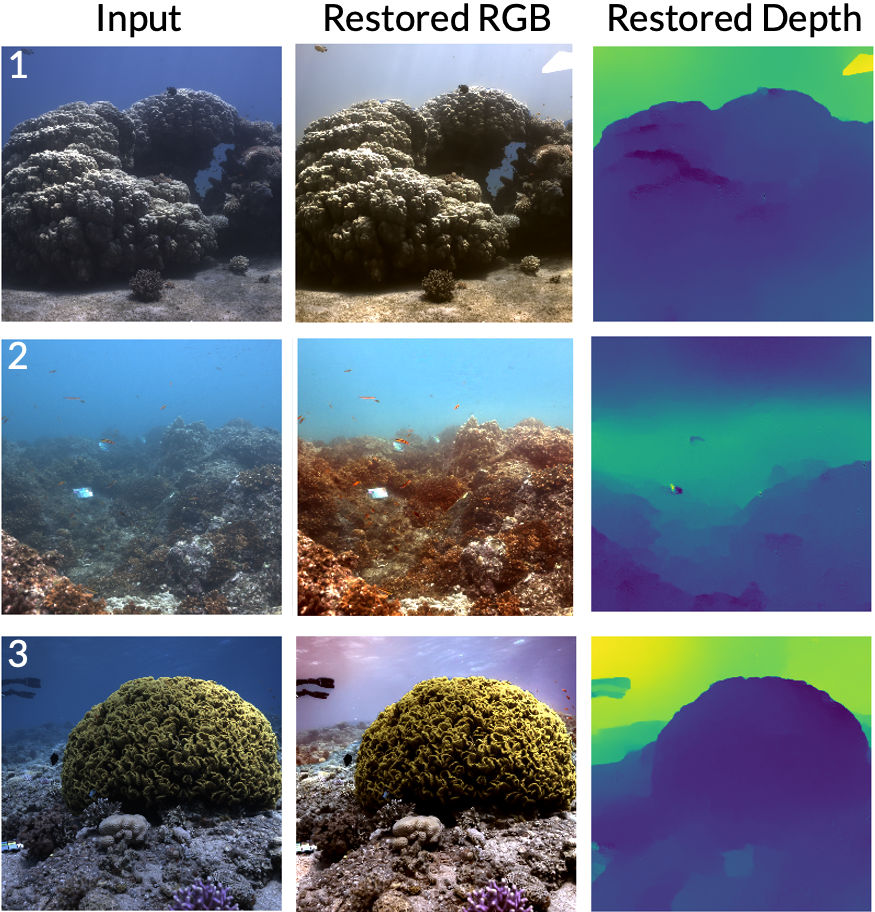}
    \caption{\textbf{Real-world failure cases.} In example~1, there is an artifact in the top-right corner. In example~2 the restored colors are reddish and saturated. In example~3 there is a pinkish hue, especially in the horizon (``sky'') area.}
    \label{fig:sup_failure_rw}
\end{figure}

\fi

\end{document}